\documentclass[10pt]{article}

\usepackage{amsmath, amsfonts, amssymb, mathrsfs, amsthm}
\usepackage{lineno}
\usepackage{hyperref}
\usepackage{stix}
\usepackage{csquotes}

\usepackage{mathtools} 
\usepackage{extarrows}

\usepackage{fancyhdr}
\modulolinenumbers[5]

\usepackage{wrapfig}

\usepackage[font={small,it}]{caption}

\usepackage{setspace} 
\usepackage{etoolbox}
\AtBeginEnvironment{quote}{\par\singlespacing\small}

\def\drawplusplus#1#2#3{\hbox to 0pt{\hbox to #1{\hfill\vrule height #3 depth
      0pt width #2\hfill\vrule height #3 depth 0pt width #2\hfill
     }}\vbox to #3{\vfill\hrule height #2 depth 0pt width
      #1 \vfill}}

\usepackage[all]{xy}
\usepackage{graphicx}

\newtheorem*{thm*}{Theorem}
\newtheorem*{mydef*}{Definition}
\newtheorem*{mylemma*}{Lemma}
\newtheorem*{myconjecture*}{Conjecture}

\begin{document}


\title{OpenCog Hyperon: \\
A Framework for AGI \\ at the Human Level and Beyond -- \\
{\it High-Level Background \& Introduction}
}

\author{Ben Goertzel, Vitaly Bogdanov, Michael Duncan, Deborah Duong, \\
Zarathustra Goertzel,  Jan Horlings, Matthew Ikle', Lucius Greg Meredith, \\
Alexey Potapov, Andre' Luiz de Senna, Hedra Seid,\\
Andres Suarez, Adam Vandervorst, Robert Werko  
\footnote{Goertzel is primary author with affiliations SingularityNET, TrueAGI, OpenCog; other co-authors are listed alphabetically } }





\maketitle

\begin{abstract}
An introduction to the OpenCog Hyperon framework for Artificiai General Intelligence is presented.   Hyperon is a new, mostly from-the-ground-up rewrite/redesign of the OpenCog AGI framework, based on similar conceptual and cognitive principles to the previous OpenCog version, but incorporating a variety of new ideas at the mathematical, software architecture and AI-algorithm level.   This review lightly summarizes 

\begin{itemize}
\item some of the history behind OpenCog and Hyperon
\item the core structures and processes underlying Hyperon as a software system
\item the integration of this software system with the SingularityNET ecosystem's decentralized infrastructure
\item the cognitive model(s) being experimentally pursued within Hyperon on the hopeful path to advanced AGI
\item the prospects seen for advanced aspects like reflective self-modification and self-improvement of the codebase
\item the tentative development roadmap and various challenges expected to be faced
\item the thinking of the Hyperon team regarding how to guide this sort of work in a beneficial direction.
\end{itemize}

\noindent and gives links and references for readers who wish to delve further into any of these aspects.
\end{abstract}



\tableofcontents

\section{Introduction}

The initial goal of the AI field, at its founding in the middle of the last century, was the creation of machines with general intelligence capability at the human level and then beyond.   When this proved more difficult than anticipated, however, the AI field shifted substantially toward ''narrow AI'' systems focused on carrying out particular tasks effectively rather than on more general-purpose adaptation and self- and world- understanding.   

With modern compute power, sensor power and data availability, though, enthusiasm for Artificial General Intelligence is resurgent and at an all time high.   It is no longer so uncommon to hear ambitious projections regarding human-level AGI potentially being achievable in some single-digit number of years.

The authors of this article have been optimistic about the outlook for AGI for much longer than this sort of attitude has been fashionable; however, we also feel that to achieve the grand goal of general intelligence at the human level and then beyond, it will be necessary to grapple a bit with the subtlety of the problem.   Human-level AGI is not the end-all of AGI by any means, but it does have quite a lot of interdependent dimensions, and we believe it will defy achievement via simplistic cognitive architectures (like current LLMs or anything centered on these).   

What we describe here is OpenCog Hyperon \cite{Hyperon2021}, a cognitive architecture and AI system design that we believe possesses the breadth and complexity to achieve AGI at the human level and beyond, via a combination of autonomous learning and human education and supervision.   Hyperon is a new system in the OpenCog lineage, founded on the same core cognitive theories and high level design concepts as the earlier ''OpenCog Classic'' system \cite{Goertzel2010h}, but redesigned from the ground up for greater scalability, usability and mathematical elegance.   We give here a high level overview of the OpenCog Hyperon system, inclusive of various theoretical and practical pursuits that have evolved around this in-progress software system and its design. 

Those who have been in the AI or AGI field for a while understand that ''branded software systems'' like Hyperon are generally not as fundamentally significant as they may appear from a technological standpoint, though they are clearly important from a sociological, marketing, or attention-gathering perspective.  Under the name ''OpenCog Hyperon'', we are bringing together diverse algorithms, data structures, mathematical ideas, cognitive system theories, and code -- which have rich interrelationship and also independent meaning, and common use in putting together, using and understanding the Hyperon software system.  Some of the discussions here will be specifically about the OpenCog Hyperon codebase and practical applications, while others will touch upon theoretical ideas and concepts that have been integrated into Hyperon from various sources, and also have significant value beyond the scope of Hyperon.

The discussion here is mostly at a high-level technical overview level.  The \url{http://hyperon.opencog.org} website contains links to various documents, videos and code repositories that delve into more specific aspects.   The reader interested in a more thorough treatment of  cognitive theory related to Hyperon is also referred to  Goertzel's 2021 paper on {\it General Theory of General Intelligence} \cite{GTGI}.

A brief note on expository style: This is an informal text put together by a group of authors who have been working together to make Hyperon a reality.   We've chosen a slightly unusual format, in which third-person prose is intermixed with direct quotes in the first person from various contributors.   The purpose of this is to make things a little less dry and impersonal, and to get across the reality that this sort of R\&D and engineering is not an abstract and faceless pursuit but rather the intersection and synergy of the work, passion and insight of a community of specific human beings, each of whom brings their own perspective and peculiarities to the project.  \footnote{ It should be noted that while {\it some} of the co-authors of this paper have been very major contributors to Hyperon (e.g. Alexey Potapov, Andre' Senna and Matt Ikle' have been. major co-creators and others have also contributed significantly), the co-author list is not a list of Hyperon creators.  Some of the co-authors have been on the periphery of the Hyperon project; and some major Hyperon contributors have not contributed directly to this article for various logistical reasons.}

\section{Snapshots Along the Path to Hyperon}

While the crux of this article is a rough description of the concrete approach to AGI being taken in the Hyperon software R\&D project, we feel it is important to situate this technical project appropriately within the broader quest of understanding ''what is a mind that we might build one?"   

Today AI and to an extent even AGI is a major commercial and practical pursuit, and a substantial portion of new entrants into the field are understandably oriented toward what they can do right now with the amazing tools that are available.   However, the OpenCog project emerged from a previous phase in the AI field, where hardware and data resources were not sufficient to support such rapid experimentation, and thus progress was largely driven by conceptual investigation -- i.e. AI R\&D was deeply bound up with the intellectual quest to understand the nature of intelligence.

\subsection{(Some Aspects Of) the Conception of Intelligence Behind Hyperon}

In a talk given at the Hyperon workshop at the AGI-23 conference in Stockholm in June 2023 \cite{Hyperon2023talk}, Hyperon co-founder Ben Goertzel reviewed some of the conceptions of general intelligence underlying his work on the Hyperon system and its predecessors.  (While these thoughts of Ben's are highly relevant, it should also be remembered that a system like Hyperon is not solely the product of one person, and there may be different perspectives on how it came about and why it exists.)

{\tt \small 'My engagement with AI started in the early 1970s, when I began reading about AI in science fiction and popular science articles. By the late 70s and early 80s, I became somewhat disillusioned with the academic field of AI, which seemed to primarily involve rule-based production systems. These systems appeared tedious and ill-conceived to me, not because I had any aversion to logic, but because the thought of manually encoding all the knowledge required for human-like cognition seemed absurd. However, as I began programming AI systems, I started to comprehend why the AI field was dominated by such systems ? achieving anything with self-organizing systems was immensely challenging.

''Another intriguing facet I discovered, in my late teens as I initially explored the AI field, was the surprising viability of narrow AI.  The existence of an AI system that could play checkers at a level superior to my own abilities seemed greatly impressive to my teenage mind, given the limited computing resources available at the time. This raised some perplexing questions about the dichotomy between narrow AI and AGI.  It was a nontrivial discovery, in the 1960s and 70s, that tasks appearing to require significant general intelligence when humans do them could often be done successfully by quite simple algorithms.  I understood why some researchers had started to think maybe human-level AGI was just a grab bag of narrow tricks carrying out various tasks that had been evolutionarily important to humans.

''On the other hand, we are now reaching a point in the evolution of the AI field where employing more AGI-oriented approaches is sometimes the most effective way to address practical problems, sometimes better than deploying simple narrow techniques even when one is operating under the severe time and resource constraints characteristic of real-world application problems.

''As I progressed through my education, getting a PhD in mathematics with a view toward understanding the mathematics of human and machine minds, I spent some time attempting to formalize the concept of general intelligence.  I settled in my 1991 book The Structure of Intelligence on the notion of general intelligence as the ability to achieve complex goals in complex environments (and then sought to formalize the relevant notions of complexity in terms of algorithmic information theory) -- a related approach to the formal definition of general intelligence given by Legg and Hutter in 2007 or so.   Narrow AIs, in this sense, tend to pursue a less complex range of goals and to operate best in contexts that are well prepared and curated by the humans or other software systems around them.

''I much later encountered Weaver -- aka David Weinbaum -- who wrote his  2018 Ph.D. thesis on open-ended intelligence at the Free University of Brussels \cite{weinbaum2017open}. Weinbaum approached the topic from a Continental philosophy perspective, examining intelligence as a complex, self-organizing, autopoietic (self-rebuilding) system. According to his view, intelligent systems are engaged in the pursuit of individuation, meaning they persist in maintaining their existence with a boundary around them, while also seeking self-transcendence by striving to evolve into something beyond themselves. He posits that achieving a goal or working towards a reward function is not the core of what makes a system intelligent. Rather, it is the self-organizing network flowing toward individuation and self-transcendence that may sometimes originate goals (implicit or explicit) and act in ways that approximate maximizing those goals. However, this self-organization can cause goals to vanish and new ones to emerge. I find that this resonates with my personal experience as a goal-oriented person. Even though I often set goals, I tend to redefine or abandon them in the process of pursuing them.  

''I found Weaver's conception of intelligence tied in AGI theory with complex systems theory in a quite satisfying, even if partly impressionistic and not fully scientifically/mathematically rigorous, way.   In the context of Weaver's concept of open-ended intelligence, achieving a goal is not the crux of intelligence. Rather, the focus is on the self-organizing aspect. For instance, an intelligent system may work towards a reward function for some time during its existence, but this is not what fundamentally makes the system intelligent. Instead, the self-organizing network, flowing towards individuation and self-transcendence, may sometimes originate goals, which may be implicit or explicit. It may act in ways that approximate maximizing those goals, but then the self-organization might cause that goal to dissipate, and lead another goal to bubble up. This dynamic nature of goal formation and the pursuit of self-transcendence felt relatable to how my own life has unfolded.

''Weaver's perspective made me reevaluate my previous fixation on achieving complex goals in complex environments, and I became more open to various viewpoints on general intelligence. It also led me to appreciate that a crisp definition is not necessary and that general intelligence may be multi-dimensional and somewhat nebulous.

''Hyperon is intended as a system that can achieve complex goals in complex environments, including goals that it represents explicitly as well as goals that arise implicitly due to its self-organizing dynamics.  It is also intended as an open-ended intelligence that, as it pursues its self-organization and its agency, ongoingly pursues both individuation and self-transcendence.''}

In the conceptual approach to AGI underlying Hyperon, human-level general intelligence is an in-some-senses arbitrary waypoint on the path to more and more generally intelligent systems.   Hyperon's basic architecture draws richly on human cognitive science, yet does not try to emulate human mind or brain in detail, and as a result is not bound to human levels of rationality, insight, analytic skill or creativity (nor bound to human levels of ethics or compassion for that matter).   General conceptions of AGI like ''achieving complex goals in complex environments'' and ''balancing individuation with self-transcendence'' do describe a lot of what humans do, but they also are clearly things at which humans are merely so-so compared to other feasible physical systems.   Making AGIs that understand and relate to human intelligence is an important thing for humanity, but Hyperon is intended as a design that will be able to achieve this alongside various forms of intelligence considerably exceeding the human level.

\subsection{What is a Mind that We Might Build One?}

Continuing to extract from the same AGI-23 talk, Ben Goertzel recounted there that:

{\tt \small
''The quest to understand the nature of general intelligence led me in my teenage years not only to AI but also the philosophy of mind, which was quite confusing but also offered some clarity in some regards.  In fact my own journey towards the OpenCog Hyperon design began more in philosophy of mind than in technical AI.

''One of the philosophers who had a significant impact on me was Charles Sanders Peirce, an American philosopher from the late 1800s. He is known for introducing quantifier logic and contributing several technical elements that found their way into AI. Peirce's MeTTaphysics evolved from three fundamental categories, which he called first, second, and third. The first is pure raw experience, which is unanalyzable. The second is reaction, such as one billiard ball hitting another, representing the physical world. The third is relationships, where one thing is relating to other things.

''These categories can be interpreted in various ways. First could be connected to raw conscious experience, as discussed by David Chalmers in his analysis of consciousness. Peirce saw the relationship between raw phenomenal experience and complex semantic relationships not as a hard problem but as a category error. His perspective was that raw experience (first), patterns of organization (third), and physical reactions (second) are separate categories that should not be reduced to one another.

"Peirce also had a notion of habits. He called it the law of the mind, which is the tendency to form habits. This meant that if a pattern had appeared in the world for a while, the probability of it appearing again would be higher than expected. The concept of patterns was fundamental to Peirce.

''Taking inspiration from Peirce, I started to conceive intelligence systems and the mind as a collection of patterns, recognizing patterns in themselves and the world, habituating to each other and randomly mutating and combining with each other as they go. These patterns, in essence, make up the mind and are seen as self-organizing systems. Other philosophers like Gregory Bateson and linguists like Benjamin Whorf had similar perspectives, viewing the world as a self-organizing pattern system.'' }

These ideas from philosophy of mind and complex systems science were fleshed out in a fair bit of detail in a series of books Goertzel released in the 1990s: {\it The Structure of Intelligence} \cite{Goertzel1993a}, {\it The Evolving Mind}\cite{Goertzel1993}, {\it Chaotic Logic} \cite{Goertzel1994}and {\it From Complexity to Creativity} \cite{Goertzel1997}.   These books also contained what could be considered high level sketches of AI designs -- but nothing concrete enough to be directly implemented without making a lot of other difficult decisions along the way.

Rather than determining an AI design, this sort of philosophical / cognitive-science /system-theoretic thinking about AGI gives a way to think about potential AGI designs or partially or wholly working AGI or proto-AGI systems, and gives guidance as to what sorts of AGI designs are more versus less likely to deliver AGI at an advanced level.

\subsection{A Partial Prehistory of Hyperon}

Goertzel's AGI-23 talk continued to explicitly recount how these conceptual investigations led to the creation of a series of practical AI software systems, culminating in Hyperon: 

{\tt \small
''In the early 1990s, I spent considerable time attempting to build systems based on self-organizing pattern recognition using early versions of Haskell. These attempts, while intriguing, did not lead to practical applications. I speculated if the issues were due to lack of scale or some fundamental problem with the approach. The human brain's sheer size, with hundreds of billions of neurons, suggested that scale might be an essential factor.

''I then discovered the potential of the internet for scaling these systems. Webmind, an initiative of the late 1990s, aimed at employing the internet to run self-organizing pattern recognition agents on a large scale.  My book Creating Internet Intelligence summarized the notion of an emergent Global Brain, and in particular the idea that powerful AGI systems (like Webmind instances) living on decentralized compute networks could serve as the cognitive hub of a centralized global brain.   

''I also in 1995 posted a Web page announcing my future run for US President on a 'decentralizationist' platform (I wouldn't turn 35 till 2001, so I wasn't old enough to run for President yet).   Creating Internet Intelligence didn't emphasize the political aspect but it was there in the background.  Well before blockchain became a thing, I and others were thinking about the need for decentralized control of the distributed AI processes comprising a large-scale AGI, and the importance of this for keeping early-stage AGIs out of the hands of centralized entities with narrow goals.   

''My early speculative designs for globally distributed Webmind systems and associated Artificial Life systems call WebWorlds combined strong encryption and distributed processing in ways strongly reminiscent of modern ledger-free blockchains like TODA.    (The notion of a huge replicated ledger of all transactions never occurred to me because of its obvious non-scalability).   However, it didn't occur to me to use decentralized money as an initial application for a global decentralized computing system -- I did think about that briefly, but it just seemed to me it would take a long time to get such a system running fast enough to compete with Mastercard, Visa and such.   Although I had done work on AI for financial trading systems, it didn't occur to me that a decentralized currency could become the center of a partially-black-market speculative trading sub-economy.  If it had, I would have been Satoshi and Hyperon would be a far better funded project!

''In any case, the technological infrastructure of the time proved inadequate for creating practical AGI systems in the form of decentralized agent systems like Webmind.  This led to the creation of the Novamente Cognition Engine, which was philosophically similar to Webmind but was organized software-wise in a configuration more akin to what Hyperon represents now. In Webmind, the focus was on a decentralized agent system, whereas Novamente was more structured. Both systems aimed at integrating logical reasoning, non-axiomatic reasoning systems, evolutionary program learning, attention allocation, and pattern mining.

''Novamente centered on a knowledge base known as Atomspace, which was a structure originally called a 'generalized hypergraph', a verbiage later tweaked to 'Metagraph.'   Basically: a graph with links that can span multiple nodes, and links that can points to links or larger subgraphs, and both nodes and links can be labeled with various more or less complex weights or structures (such as types that come from complex type systems and may be represented as Metagraphs themselves).   Nodes and links in this framework are both referred to as Atoms.   Webmind contained all this representational capability as well, but organized in a less clear and elegant way.

''The Novamente team -- led by myself, Cassio Pennachin, Andre' Senna (who now plays a lead role in Hyperon), Pei Wang (whose NARS AGI initiative is still moving forward impressively) and others --  implemented various cognitive processes as agents acting on this Atomspace. Essentially, we restructured the nesting order of two loops. In the original structure, the outer loop iterated through agents, each of which enacted different functions that collectively carried out cognitive processes. With Novamente, the looping sequence was reversed, as the system now looped through cognitive processes that acted on nodes and links within the Atomspace. 

''The efficiency of either approach was dependent on the hardware infrastructure in use: On an architecture like Danny Hills's Connection Machine, a MIMD-parallel hardware device with up to 128000 fully independently programmable processors which I experimented with briefly in the mid-1990s, a Webmind-like design would be maximally appropriate and effective.   On typical contemporary computing systems with their strict 'von Neumann architecture' separation between RAM and processing, a design more like Novamente has more hope of achieving acceptable efficiency.   On more innovative modern hardware designs like the processor-in-RAM AGI chip we're working on with Simuli and TrueAGI, the optimal approach to implementing this general sort of AGI system may be somewhere between Webmind and Novamente, in a sense.  

''The Hyperon framework, unlike either the Webmind or Novamente systems, is designed to support a lot of flexibility in terms of how fully 'decentralized-agenty' an implementation is -- which is possible because of a lot of advances in computer science and software design over the last two decades.   This ties in of course with the use of Hyperon's MeTTa programming language (whose programs are subnetworks of Atomspace, designed to be interpreted as programs for transforming Atomspace) as a smart contract language in the Hypercycle ledgerless blockchain, which we'll talk about a little later.

''The Novamente AGI initiative was successful in producing research papers exploring different AGI algorithms and achieving some limited practical deployments, especially in natural language processing and signals analysis. In 2008, significant portions of the system were open-sourced as OpenCog, and a community of contributors began to build upon this codebase. One significant contribution was the OpenCog Pattern Matcher, developed by Linas Vepstas. Initially, the Pattern Matcher was a system for recognizing patterns in the Atomspace knowledge graph. Eventually, it evolved into a functional and logic programming framework with recursive pattern matching capabilities.

''However, as OpenCog evolved, certain limitations were encountered, primarily related to scalability and ease of use. To address these challenges, some of the developers began experimenting with neurosymbolic systems. Alexey Potapov and Vitaly Bogdanov interfaced Torch, a deep learning library, with OpenCog, which allowed for combining symbolic and neural processes. However, this integration was found to be somewhat inefficient, as OpenCog was considerably slower than modern neural network frameworks, especially when running on GPUs.

''Which is a meaningful part of what led us toward the AGI infrastructure design now referred to as OpenCog Hyperon.''}

The need for a more scalable and usable system than OpenCog circa 2020 led a number of OpenCog developers to consider rebuilding the system almost from scratch. This was coupled with ongoing developments in mathematics and theory which these developers wanted to incorporate more directly into the AI system. One such concept was cognitive synergy, which stemmed from an understanding of how different types of memory and learning mechanisms in the human brain interact. Cognitive synergy highlighted how the brain translates a problem from one kind of memory to another kind when it gets stuck, essentially employing different kinds of learning mechanisms.

To formalize this interaction between declarative, procedural, and sensory knowledge, category theory and other related mathematical concepts were used to map various AI algorithms into operations over a metagraph.  It began to seem feasible to incorporate this mathematical framework more deeply into OpenCog to make it more efficient.   This conceptual and formal direction also seemed promising in terms of making more elegant and efficient interfaces between OpenCog systems and external AI systems like scalable deep learning frameworks.

Eventually, based on these needs and inspirations, a new system called OpenCog Hyperon was developed.   'Hyperon' is the name of an elementary particle; the name was basically chosen to continue the physics metaphor theme originated in 'Atomspace', and for its resonance with 'hypergraph' (since Atomspace is a generalized hyper graph aka Metagraph).

Tongue partly in cheek, it was proposed that the next huge overhaul might be called OpenCog Tachyon, incorporated quantum computing constructs at the core, or maybe computational acceleration using novel hardware incorporating closed timelike loops!  (But we'll leave that to a later document...)

\section{OpenCog Hyperon: A Modern, Scalable Infrastructure for AGI}

Now let us finally go into a little detail about Hyperon as a software framework!  

\begin{figure*}
\begin{centering}
\includegraphics[width=15cm]{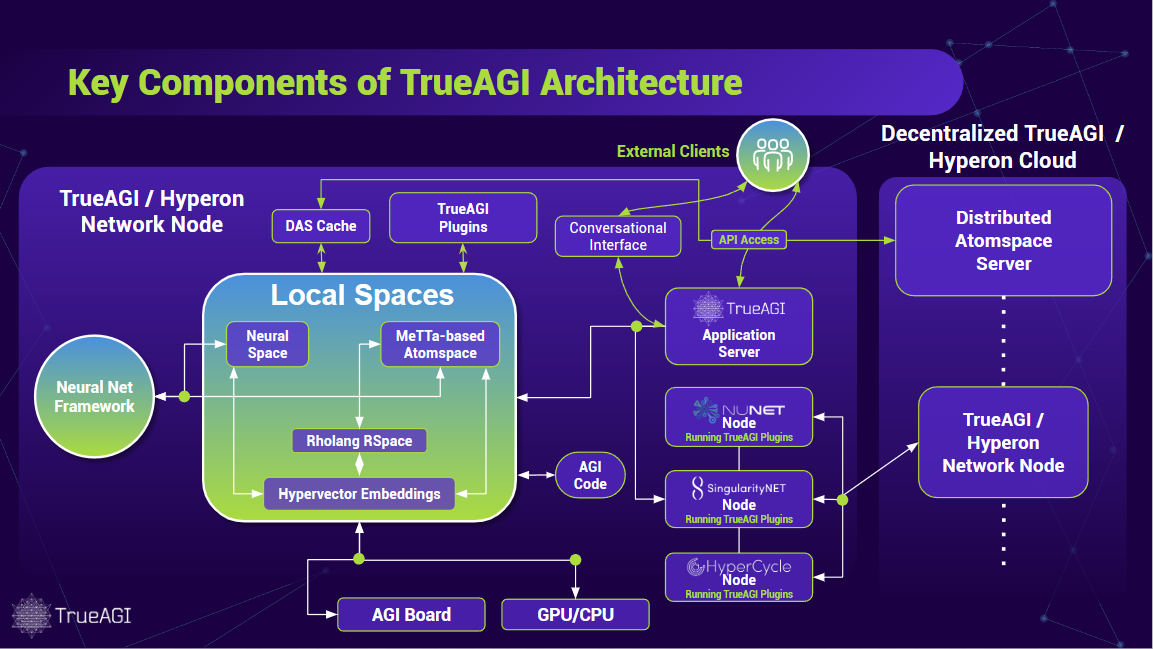}
\protect\caption{\label{fig:hyperon} High-level illustration of key components in Hyperon architecture, including integration into TrueAGI application framework.}
\end{centering}
\end{figure*}

Figure \ref{fig:hyperon} gives a rough overall depiction of many of the key components, though the interrelation of these components is highly diverse and dynamic and defies simple accurate diagrammatization.

\subsection{Atomspace and MeTTa : The Core Constructs of OpenCog}

The core construct of Hyperon, like that of OpenCog and Novamente Cognition Engine before it, is the Atomspace, a metagraph comprised of nodes and links with complex interlinkage structures. This metagraph is highly versatile and allows for labeling nodes and links with various kinds of data, including subgraphs. This labeling mechanism can also facilitate the embedding of complex type systems in the metagraph.

A new ingredient of Hyperon, qualitatively different from any thing in earlier OpenCog versions, is the programming language called MeTTa.   MeTTa programs are sub-metagraphs in Atomspace, and are interpretable as procedures for rewriting portions of Atomspace into modified or additional portions of Atomspace.  (In a trivial way, every sub-metagraph in Atomspace is a sort of MeTTa program, in the sense that every Atom that's not interpretable as some instruction by the MeTTa interpreter is still a ''constant'' items of data processable by MeTTA programs.)

The prior version of OpenCog, which in the Hyperon era has come to be called 'OpenCog Classic', featured a sophisticated pattern matcher (designed and implemented by the heroic OpenCog developer Linas Vepstas) which had added into it a variety of sophisticated functions allowing it to do more than just match patterns in a traditional sense, but also execute various transformations on the Atomspace while in the course of doing pattern matching.   This is a reasonable and interesting design, but in Hyperon it was decided to do things a bit differently, and to instead:

\begin{itemize}
\item Make the Hyperon pattern matcher more of a standard, static pattern matcher (though with plenty of special aspects, such as subtle handling of patterns involving variables, efficient execution of matching involving Atoms belonging to type systems with efficient type inference associated with them, and matching of variables against whole subgraphs rather than individual Atoms)
\item Instead of embedding complex programmatic logic in the pattern matcher, create a language MeTTa wrapped around the pattern matcher, which has invocation of pattern matching as a key functionality, but does programmatic logic outside rather than within the traversal process of the pattern matcher
\end{itemize}

MeTTa then becomes a quite novel sort of programming language incorporating aspects of both functional and logic programming, but going beyond either of these standard paradigms.

Alongside pattern matching, MeTTA also includes equalities with a distinct semantics that took substantial consideration to develop.  One aim here was to integrate MeTTa with homotopy type theory among other mathematical models that provide complexity to the concept of identity.   MeTTa's unorthodox approach to dealing with equality is low-level enough to not commit too much to how assignment or comparison have to operate. 

By its basic nature, the MeTTa interpreter enables the Atomspace to undergo a metamorphosis and initiate self-rewriting. This has obvious implications for self-modifying code. The Atomspace, when coupled with the MeTTa interpreter, can essentially be perceived as self-modifying and self-rewriting sets of nodes and links.

Additionally, the MeTTa language is extremely versatile and can be represented as both a metagraph and a MeTTa-graph. 

As Alexey Potapov recounts, in the early phases of Hyperon design it was considered whether to use existing graph or vector stores and associated query engines rather than build our own new Atomspace and associated tools, but in the end the requirements for AGI were just too distinctive:

{\tt \small "Representation of knowledge in Hyperon including MeTTA programs is metagraphs, which constitutes a substantial practical distinction from most other AI and AGI approaches out there.  While ordinary graphs can be described as a collection of triples, and hypergraphs are collections of tuples, metagraphs are collections of trees, i.e., each edge is a tree-like connection of nodes or, alternatively, each edge is a tuple connecting any number of nodes and other edges. This representation is crucial for representing complex statements and arbitrary knowledge, and it is also convenient to represent program code. While metagraphs can be encoded in simpler data structures like ordinary graphs, their traversal, indexing and retrieval algorithms are not optimized for metagraph encodings. The latter requires introducing auxiliary nodes, which should be treated specially in indexing and traversal. While tweaking graph databases in such a way is possible in principle, benefits of the underlying graphical representation and corresponding algorithms are not obvious.
 
"Moreover, the very core operation of MeTTa is pattern matching with unification of variables both in the query and knowledge base entries to be matched. This feature is supported in none of the query engines, and its implementation is achievable in a less cumbersome way for lower-level host representations like key-value storage. This is the reason for not using graphical storages as back-ends both for in-RAM Atomspace and Distributed Atomspace."
}

Greg Meredith, in collaboration with the core MeTTa developers Alexey Potapov and Vitaly Bogdanov along with Adam Vandervorst, has put together a formal operational semantics for MeTTa which can be found here\cite{meredith2023metametta}.   There is also a Formal MeTTa code repository \footnote{\url{https://github.com/Adam-Vandervorst/FormalMeTTa}} which sticks very close to the operational semantics, and has been used to prototype ideas before porting them to the main MeTTa codebase.

The relation between MeTTa and type systems, as used in computer science and formal logic, is worth of brief mention even at this overview level.   MeTTa is extremely general at its core. It does not possess types as it is fundamentally just rewrite rules on the metagraph that are embedded within the metagraph. However, since types are merely portions of the metagraph that are attached to selected nodes and links, type systems can be built within it. This results in an extremely generalized version of type systems that can handle dependent types, gradual types, higher order types representing probabilities, and so forth. The challenge remains in writing efficient type checkers for these elaborate types, which involves intricate computer science.

As an illustration of the elegance and generality of this programming framework, at the AGI-22 OpenCog workshop, Jonathan Warrell demonstrated a very concise implementation of Aczel's non-well-founded set theory in MeTTa \cite{warrell2022meta}. The sets in this theory can contain themselves as elements or can have cyclical containment. This implementation made use of circular graphs to represent non-well-founded sets in the Atomspace, establishing rules to rewrite them.   This allows among other things an elegant implementation of probabilistic reasoning on infinite-order probabilities, which Goertzel has proposed as a way to handle uncertain shared social knowledge (''we both have intuitive knowledge that we both know this statement is probably true'' and other more useful variations like the shared understanding that characterizes every I-Thou relationship and every robust culture).

There are close connections between Stephen Wolfram's concept of the Ruliad, pursued from a theoretical physics perspective, and MeTTa's Atomspace. Both are meta-graph structures representing patterns among patterns among patterns ... (without any recursion limit) ... and both can be modeled using infinity groupoids and infinity comma one categories, among other related mathematical constructs.  The specific portions of the Ruliad Wolfram has been interested to explore in his physics-inspired work is however different from the specific sorts of metagraphs that appear most useful in an AGI context in the near term.

As regards the name MeTTa, it is an acronym for Meta-Type-Talk, but as Ben Goertzel noted in his AGI-23 talk, {\tt \small ''It's also not escaped our notice that the term metta, in Buddhist philosophy, signifies loving-kindness.  Impressionistically, the name could be interpreted as guiding the system towards a benevolent AI, both philosophically and in computer science terms.''}

\subsubsection{The Necessity of One More Programming Language}

As Ben Goertzel noted in his AGI-23 talk. {\tt \small "After some early attempts at creating novel AGI languages in the mid-1990s, I came to the conclusion that creating new languages was a temptation that the serious AGI researcher should probably resist.   So often, it seemed, a researcher had the idea that if they just had a more apropos programming language their work implementing AGI would be easier -- and then 30 years later they looked up and found most of their career had been spent on programming language R\&D not AGI.....  But yet, here I am several decades into my AGI research career, helping create a new programming language.   A difference from some of the historical examples like LISP or Prolog, though, is that in the Hyperon team we are coming at our new language from the perspective of having quite a lot of specific stuff that we want to implement in the language.   We already have built a bunch of specific prototype proto-AGI systems in other languages and frameworks, so we have a pretty good idea of what properties and aspects we need a language to have in order for it be maximally usable  to realize our AGI ideas effectively."}

Alexey Potapov reiterates these points in more detail: {\tt \small "After long consideration of alternatives, we decided to embody knowledge representation, pattern matching queries, and their chaining in a dedicated language, MeTTa, which is the fundamental component of Hyperon. The need for a representation of knowledge is obvious, and introduction of a cognitive language is also very typical for cognitive architectures and similar platforms, but we did consider long and hard if it should be a separate programming language? Cognitive languages can be rather restricted or specialized and can be implemented as DSLs or libraries in some general-purpose programming language.
 
"Since Hyperon is a platform for cross-paradigmatic studies, it was clear its cognitive language should not be too specialized.
 
"Such languages as Agda, which serve as proof assistants, or more classical logical language like Prolog, as well as universal probabilistic programming languages such as Church or Anglican, are focused on one paradigm. For example, Prolog does have such extensions as ProbLog, but they are stand-alone projects, which are not interoperable. Similarly, when the need arises to analyze execution traces of probabilistic programs, one needs to hack into the language interpreter. Such languages do not natively work with knowledge bases, but can may only query external graph databases, implying that they do not have many features of cognitive architectures.
 
"Implementing Hyperon as a library in some general-purpose host language also encounters some obstacles. As was mentioned above, such use cases for Hyperon exist, in which all knowledge entries including reasoning rules and procedural knowledge should be introspectable and rewritable. If we imagine Hyperon as a library in, say Python like PyTorch, we would like to write as much as possible in the Hyperon language, which programs are knowledge in the Atomspace, and as little as possible in Python. Code in Python is not introspectable knowledge, which can be reasoned over. Many examples of ad hoc symbolic systems implemented in traditional languages exist, but processing of symbolic information in such systems is written in imperative languages, which make them incompatible with each other and not usable by a larger symbolic system. Implementation of inference rules in the cognitive language itself is thus very important. This, however requires, such language to be quite general and rich. While it doesn't mean that this cognitive language should be as general-purpose as Python or Rust implying that it should benefit a lot from interfacing with general-purpose language, this means that this cognitive language should be enough to implement various AI systems and components. Being put into Atomspace, these programs will not be represented as the code in some host language, and will be evaluated by their own interpreter. While it is still possible to avoid having one's own syntax for this cognitive language and fill in Atomspace using constructions in some existing host language, abstracting the host language syntax away and separating pure cognitive language code have additional benefits.
 
"Basically, Hyperon, similar to OpenCog Classic and many other cognitive architectures, has its internal cognitive language, which differs only in its universality and flexibility due to the design of Hyperon in general. Thus, it is not an option to not have any language at all. And whether it is treated as a programming language or ''just'' an internal cognitive language without possibilities to write libraries in it, debugging, etc. is the matter of convenience. Apparently, if we want to implement different components like PLN, MOSES, ECAN, etc. for Hyperon in MeTTa as well as other libraries with custom inference strategies and different ways of integration with DNNs and such, then it is more convenient to treat MeTTa as a programming language."}

\subsubsection{MeTTa in a Nutshell}

MeTTa is a novel language and getting used to it requires letting go of some traditional ways of thinking about programming, and acquisition of some new habits ... no quick overview is going to provide a fully effective shortcut in this regard.   However, for readers who have sufficient technical background, it seems a brief summary of highlights and differentiating aspects may still be of some value!\footnote{This section was written by Alexey Potapov}

Programs in MeTTa are collections of expressions, which are placed into a container called Atomspace. Expressions are tuples of atoms, which can be either other expressions, or pure symbols, or grounded atoms, entities that wrap some (subsymbolic, not fully described by the MeTTa program itself) data. For example, \verb|MySymbol |, \verb|(A (B C D) E)| and \verb|(''point'' (10 10))| are valid expressions, where \verb|''point''| and \verb|10| are supposed to be turned into grounded atoms by the parser.
 
Atomspaces differ from other containers in that they support special types of queries for retrieving expressions from them. While MeTTa expressions can treated as edges in the metagraph, and the whole Atomspace can be treated as a metagraph data/knowledge base with a querying engine efficiently operating over an indexed content, Atomspace queries can be understood also as a generalization of pattern-matching in functional languages. Queries are expressions typically containing a special type of atoms, variables (in the current syntax they are distinguished from ordinary symbols by placing \verb|$| at the beginning of their name). Expressions with variables can be referred to as patterns. For example, expression \verb|(A (B C D) A)| can be retrieved by queries (or matched against patterns) \verb|(A $x A)| or \verb|(A (B $x $y) A)| or \verb|($x (B C D) $x)|, but not by \verb|(A ($x $y C) A)| or \verb|(A (B C D) (A $x))| or \verb|($x ($x C D) A)|.
 
The difference between Atomspaces and many other database containers is that they can contain expressions with variables, and queries are matched with such expressions if variables both in queries and in expressions can be bound to some subexpressions in a non-contradictory way. For example, \verb|(A ($a $a) A)| can be matched against \verb|($b (B B) $b)|, but cannot be matched against \verb|($b (B B) C)| or \verb|($b (B $b) $b)|. In the latter cases, we cannot find such a substitution for both \verb|$a| and \verb|b|, which will make these two expressions identical.
 
The core pattern matching function receives a query pattern and a result pattern as its parameters. It searches for expressions in the specified Atomspace, which can be unified with the query pattern, and outputs the result pattern substituting variables in it with the values found during unification. For example, if the query pattern is \verb|(A (B $x $y) A)| and the resulting pattern is \verb|(Found $x $y)|, and the query pattern is matched against \verb|(A (B C D) A)|, then the result will be \verb|(Found C D)|.
 
Some grounded atoms can wrap executable code in the host language. MeTTa expressions can be evaluated, and if an expression starts with an executable grounded atom, evaluation of this expression will result in execution of the wrapped code.  \verb|match| is a grounded atom, which refers to the implementation of pattern matching \verb|&self| is a grounded atom which refers to the Atomspace in which the MeTTa program itself is stored. Evaluating 

\begin{verbatim}
(match &self $(A (B $x $y) A) (Found $x $y)) 
\end{verbatim}

\noindent as an expression in MeTTa will result in a call to the pattern matching over the program Atomspace with the corresponding query and result patterns.
 
When a MeTTa script is processed, its expressions are put into the program Atomspace. If one wants to evaluate an expression immediately, currently, ! should be put before this expression. In the course of execution of the following MeTTa program

\begin{verbatim}
(Sam is a frog)
(Tom is a cat)
(Sophia is a robot)
! (match &self ($x is a robot) (I know $x the robot))
\end{verbatim}

\noindent will put the first three expressions in the program Atomspace, and evaluation of the last expression will produce \verb|(I know Sophia the robot)|.
 
Patterns in the program Atomspace are useful for representing some general knowledge. The following program

\begin{verbatim}
(Implies (Human $x) (Mortal $x))
! (match &self (Implies (Human Socrates) $y) (Concluding $y)))
\end{verbatim}

\noindent will output \verb|Concluding (Mortal Socrates))|, because the query can be unified with the Atomspace expression, when \verb|$x| is replaced with Socrates, and \verb|$y| is replaced with \verb|(Mortal Socrates)|.
 
While evaluation of grounded operations is delegated to executable code wrapped by them, evaluation of symbolic expressions is performed by the interpreter constructing equality queries. That is, if \verb|$(f a)|, is evaluated, query \verb|(match &self (= (f a) \$r) \$r)| is constructed, and the result of this query is evaluated further. If the result of some equality query is empty, the expression is evaluated to itself (not reduced). Let?s consider the following program:

\begin{verbatim}
(= (add (S $x) $y) (Add $x (S $y)))
(= (add Z $x) $x)
! (add (S Z) (S Z))
\end{verbatim}

Evaluation of \verb|(add (S Z) (S Z))| will start with the query pattern \verb| (= (add (S Z) (S Z)) $r)|. It can be unified with the first expression in the Atomspace with bindings: \verb|$x<-Z, $y<-(S Z)|, and \verb|$r<-(add Z (S (S Z)))|. \verb|$r| is interpreted further by constructing the query pattern \verb|(= (add Z (S (S Z))) $r)|. It will be unified with \verb|(= (add Z $x) $x)| yielding \verb|(S (S Z)))| as the result. Attempting to query \verb|(= (S (S Z)) $r)| further will give an empty result \verb|(no match)|, and \verb|(S (S Z)))| will be evaluated to itself (the final result).
 
It can be seen that evaluation of MeTTa expressions via equality query chaining works as functional programming. But one should note that these equalities are still entries in the knowledge base, which can be explicitly queried. For example, one can execute query 

\begin{verbatim}
(match &self (= (add $x $y) Z) (Answer $x $y)) 
\end{verbatim}

\noindent for the last program and get \verb|(Answer Z Z)|, because the query pattern can be unified only with \verb|(= (add Z $x) $x)| (note that variables with the same names in the two patterns to be matched are treated as different variables).

Queries with variables are similar to database queries. Program expressions with variables are similar to functional programming. Variables on both sides enable features similar to logic programming (united with functional programming and knowledge bases in a unified way). Consider the following example (\verb|True| and \verb|and| are defined in the standard library):

\begin{verbatim}
(= (croaks Fritz) True)
(= (eat_flies Fritz) True)
(= (frog $x)
   (and (croaks $x)
    	(eat_flies $x)))
(= (green $x)
   (frog $x))
! (green Fritz)
\end{verbatim}

\noindent The last expression will be evaluated to \verb|True| in a functional way. At the same time, \verb|(green Sam)| will not return \verb|False|, but it will be reduced to \verb|(and (croaks Sam) (eat_flies Sam))|. More interestingly, \verb|(if (green $x) $x (no-answer))| can also be evaluated producing Fritz. The reason is that MeTTa can evaluate expressions with variables, because it simply constructs equality queries, e.g. \verb|green $x)| is evaluated via the query pattern \verb|(= (green $x) $result)|, \verb|$result| will be \verb|(frog $x)|, which will be evaluated further to \verb|(and (croaks $x) (eat_flies $x))|. Query pattern \verb|(= (croaks $x) $result)| will be matched against \verb|(= (croaks Fritz) True)|, with \verb|$result| bound to \verb|True| and \verb|$x| to \verb|Fritz|.
 
With the use of \verb|match|, programmers can define rules for inference over purely declarative knowledge in MeTTa if they want to avoid automatic equality-based chaining. For example, the program \verb|(,| in the query pattern means that two subpatterns should be matched simultaneously with possibly different expressions, but with same variable bindings)

\begin{verbatim}
(Fact (Human Plato))
(Implies (Human $x) (Mortal $x))
! (match &self (, (Implies $a $b) (Fact $a)) (Inferred $b))
will output (Inferred (Mortal Plato)).
\end{verbatim}
 
MeTTa has a number of specific features:

\begin{itemize}
\item Non-determinism. Queries can return multiple results, and thus equality-based evaluations can return multiple results as well. \verb|(match &self (is-a $x Human) $x)| will return both \verb|Plato| and \verb|Socrates| if 
\begin{verbatim}
&self contains (is-a Plato Human) and (is-a Socrates Human)
\end{verbatim}
\item Gradual dependent types. Symbols can be typed, and types of expressions will be automatically inferred (via pattern matching as well), e.g., for
\begin{verbatim}
(: Nat Type)
(: Z Nat)
(: S (-> Nat Nat))
(: Vec (-> $t Nat Type))
(: Cons (-> $t (Vec $t $x) (Vec $t (S $x))))
(: Nil (Vec $t Z))
\end{verbatim}
\noindent the type of \verb|(Cons 0 (Cons 1 Nil))| will be \verb|(Vec Number (S (S Z)))|.
\item Custom grounded atoms. MeTTa programs can be extended by grounded atoms wrapping custom external data structures or code provided in another language (Rust, C++, Python). This is an important feature for grounded reasoning and neural-symbolic integration.
\item Self-modification. Besides match, Atomspace API includes functions for adding and removing atoms, which themselves are represented as grounded atoms in MeTTa, so the program in MeTTa can completely rewrite its own code.
\end{itemize}

\subsubsection{Optimizing the MeTTa Interpreter for AGI}

MeTTa as a language could be used for a variety of applications beyond AGI -- there are many different domains where having a language flexibly capable of both logical and functional programming, and able to host a variety of different type-systems including those created at run-time, would simplify otherwise complex programming tasks.   However, our. main goal with MeTTa is precisely AGI, and so we have put significant analysis into how to optimize the MeTTa interpreter for effective performance on the array of AGI algorithms we consider likely to be most critical.

As Ben Goertzel notes, 

{\tt \small "In 2020 I found myself with more time for theoretical work than in the years immediately previous, due to the COVID-19 pandemic massively slowing down my business travel schedule.   So I revisited some mathematical AGI theory that I'd set aside some years before, and did some work convincing myself that basically all the algorithms needed for AGI according to my OpenCog approach could be represented, within a decent degree of approximation, via fancy versions of operations called folding and unfolding, enacted over metagraphs.

"In particular, there are operations in functional programming theory called futumorphism, histomorphism, metamorphism and so forth -- the so-called 'morphism zoo' -- which are usually implemented over lists or trees, but it seemed to me that by implementing these 'recursion schemes' over metagraphs one would have an infrastructure for efficiently implementing logical reasoning, evolutionary learning, attention allocation, and so forth ... all the core techniques I felt were critical for implementing human-like mind.

"The implication here is, then, if one can make these folding and unfolding operations on metagraphs efficient in the MeTTa interpreter, this will go a long way toward making AGI algorithms efficient in the Hyperon system.   I.e. there is value in doing some math to reveal more of the common structures and operations underlying what appear to be quite diverse AGI oriented algorithms.    And while this math is not incredibly abstract or difficult compared to the toughest modern mathematics, it's also stuff that basically couldn't have been done 20 years ago, because not enough of the math of functional programming had been developed at that point.

"A disadvantage of this sort of approach is that it's a bit too abstract for the typical computer programmer to deal with, let along say an application developer or a cognitive scientist.  But of course few users of a computer or a phone need to understand the underlying semiconductor physics either.    This becomes a challenge in system design -- creating domain-specific languages, APIs and other simplified tools that enable usage of sophisticated underlying mechanisms without needing understanding of all the details. "}

\subsubsection{Some Challenges for the Next Phase of MeTTa Development}

MeTTa is a quite different sort of programming language from anything else out there, and its closest relatives are relatively obscure languages known only to hardcore functional programming aficionados.   For those who really want to understand how it works in practice, some of the walk-through videos given by Alexey Potapov would be a good place to start \cite{potapov2022mettaDSL} \cite{potapov2022metta}, along with Alexey's basic conceptual documentation written at the start of the MeTTA project \cite{potapov2021metta}.   For those with a more strongly mathematical bent, Greg Meredith's write-up of the MeTTa operational semantics will shed some clarity \cite{meredith2023metametta}.

While a lot has been done to get MeTTa to the point where it is now -- and the language is already being used for some experimentation with AGI-oriented algorithms -- there is a lot of work needed to progress MeTTa to the point it needs to be in order to serve effectively as the underlying language for development of and reflective self-programming of a human-level (and beyond) AGI.   Readers who have gone through some of the above in-depth materials on MeTTa may be interested in Adam Vandervorst's comments on some of the challenges faced in the next phase of MeTTa development.  Adam's Formal Metta codebase is also interesting to look at, and has been very useful as a sort of intermediary between the purely mathematical view of MeTTa and the primary Rust codebase that has been the focus of proto-AGI experimentation.

Adam summarizes some of his thoughts on the present and future of MeTTa as follows:

{\tt \small
At the core of the Hyperon ecosystem lies the Atomspace, the spiritual successor of the OpenCog Atomspace, built from the ground up to support the new MeTTa language (in turn the successor to the Atomese language). Both are radically simplified: no first-class lambda expressions or quotes, and a type system that lives at the same level as the terms. Manual query building has been replaced with a uniform (nested) unification syntax, and adding custom adding functionality has been fully embraced. The simplifications allow for many of the advancements in this document, and we'll now discuss some of the challenges they bring.

\paragraph{Simple term language}: MeTTa's bare-bones term language provides great power and flexibility but to make it really usable by a wider developer group there will need to be a lot more documentation and tooling.

MeTTa has just four types at its core: symbols (names), variables (placeholders), grounded symbols (custom functionality), and expressions (lists of these core types). This means data structures like Algebraic Data Types have to be declared and used in the same way as any other construct. The same is true for functions, which have to be broken up in scopes where variables can be unified, and then noted down as any old expression with variables. Let's take a look at this more concretely:

\begin{verbatim}
(= (succ \$x) (S \$x))
\end{verbatim}

\noindent That doesn't look too bad of a definition! None of these symbols are grounded, we have one variable in \$x and the rest are symbols. However, some renaming gets us here:

\begin{verbatim}
(If (Has $money) (make $money))
\end{verbatim}

\noindent which loses all intent of the original statement. Generality cuts both ways, and not having a ?new? keyword for instantiating classes or a ''def'' keyword for defining function causes the language to be extremely flexible. Even more so considering renaming statements (equalities of their own) can be added at runtime.

Not all hope is lost at creating a friendly and safe language, though. Recently, languages have been separating naming from functional definitions. Unison is one such language which stores code in a database, and keeps separate files naming the normalized code fragments. Let's take a look at Lamdu \footnote{\url{https://www.lamdu.org/}}: they embrace the naming flexibility to provide an unseen level of internationalization. The interactive nature of their editor allows users to see what names are bound to by virtue of virtual inlining, live evaluation, and showing aliases on hover. Tooling will be paramount in the success of a low level, express, and dynamic language like MeTTa.

Note how in our example, we broke expectation by:

\begin{itemize}
\item Aliasing the symbol `=` which plugs in directly into the interpreter via match and evaluation.
\item Aliasing the function `succ` by a property or destructor looking `Has`.
\item Aliasing the constructor `S` by a function looking `make`.
\end{itemize}

In an editor functions can be rendered as such regardless of the symbol indicating it plugs into evaluation. Newly defined names can be highlighted like is conventional in programming. References can be resolved by the editor, to indicate whether a symbol has a definition in the space or not.

All these solutions make use of the dynamic database nature that brought about the challenges in the first place. This puts extra burden on the library and editor authors: define in MeTTa itself what symbols in MeTTa statements mean, so it can be queried by the user (their tools): quite a meta solution. 

\paragraph{Extensible type system }  The base MeTTa language is untyped, though complex type systems can be created on top of this core, which is anticipated to be a major usage mode of the language.  The untyped feel of the base language has its pros and cons.

The unification and transformation MeTTa supports is enough to implement type checking, inference, and elaboration. However, because MeTTa programs are long lived and built over time rather than space, ideally you're engaging in a conversation with type checker, rather than committing your entire program at once and getting a report of type errors back. Together with types being just terms, e.g.

\begin{verbatim}
(: succ (-> Nat Nat))
\end{verbatim}

\noindent You're optionally narrowing down the valid terms as you're building the program. On the positive side, this allows you to experiment, either defining the type or the term first. Potentially, with an editor, even allowing you to insert the type guessed by the typer in the code base. On the flip side, you have to keep the level of completion of your program in your mental model. That is, the interactive process won't complain about missing definitions, and will instead execute to where it can. So you and your tooling are responsible for planning and marking the todo's of your program, not unlike proof systems where you have ??? indicating holes and the type elaborator helping you fill them.

\paragraph{Simple expressive reduction semantics}   For computer scientists, functional programmers or AI developers accustomed to languages varying on the lambda calculus formalism, the lack of explicit lambda-like mechanisms in MeTTa may seem counterintuitive.

Instead of the classical beta reduction 'variable substitution', MeTTa uses unification. This two-sided substitution mechanism, also seen in the new Verse programming language (and its calculus) \footnote{ \url{https://github.com/UnrealVerseGuru/VerseProgrammingLanguage}}, allows information to flow two ways: not only can the results be computed from the arguments, the arguments can also be computed from the results, and any combination of those. This is quite alien for folks coming from almost any other paradigm than functional logic programming, but it allows (locally) declarative statements you mathematically expect to work to be used in programming.

Unification is a heavy hammer, though. In many contexts you don't need it, and you have to reason about the potential meaning for longer because the possibility of 'odd things' happening is always present in MeTTa. This can be tackled in a similar fashion to renaming (supra) and grounding (infra) where extra information, preferably statements in MeTTa, can indicate the control flow to the user and optimizer.

\paragraph{Transforming spaces with match} As alluded above, MeTTa has a unified construct for all your querying needs.   The 'match' construct is used to handle anything from the most benign lookup -- e.g.

\begin{verbatim}
sensors[temperature][outside]): 
(match &sensors (Temperature Outside $t) $t)
\end{verbatim}

\noindent -- to two-step transformations (classically an orchestration of joins and meets on different fields chained into insertions) such as

\begin{verbatim}
(match &process (send $channel, $payload) 
(match &process (recv $payload $channel \$body) 
 (Result \$body)))
\end{verbatim}
 
\noindent Note that without an extra (JIT) compilation step a database query is executed on every result of the first database query.

This sort of  optimization problem (that also applies to functions on multivalued inputs) has been tackled by many groups before, though unification adds an extra layer of complexity. 

\paragraph{Custom functions} The potential to use 'grounded Atoms' to refer to functions implemented in other programming languages outside the Atomspace is critical and valuable, and can be used quite flexibly, but also presents the developer with complex choices.  

The questions 'Is this grounded? Should this be grounded?' don't always have obvious or single answers.

From basic functionality like floating point numbers to whole PyTorch models, everything outside of the four core types is added via grounding. Grounding can change the value, pattern matching behavior, and evaluation. Since it's outside of MeTTa, and the grounded definition is not queryable like a normal definition, the library author has the responsibility to add an elaborate description. For example, if the definition is in MeTTa, we can do a query to find out the number of arguments a function takes, but if it's a grounded function, we need to either add this information to a meta-space or use a stub.

A more complicated matter is dependence. A `match` (and evaluation in general) takes into account specific symbols. For example, `=` is not a grounded symbol, but it is relied on in a grounded function. If `match` was implemented in MeTTa, we could write a function that analyzes the definition and informs us about aliasing `=` (or writing a grounding for it ourselves). For example, a developer relied on the symbol `,` to mean pair, but at some point the grounding of `match` was changed to give special meaning to this symbol (using it to indicate a composite query), which broke the developer's code.

There are several potential solutions, from namespaces, to code sanitizers and interpreter warnings. Coming up with a holistic solution will require significant effort, and is needed to scale MeTTa to serve large codebases and long dependency chains.
}

\subsection{Hyperon's Position in the Era of LLMs}

As one can see from the last few sections of this document, MeTTa and other aspects of Hyperon have their own considerable technical depth, calling on a variety of academic and practical knowledge bases and traditions with their own long histories.   On the other hand, today's AI scene is somewhat heavily dominated by LLMs, which in some ways are coming from a quite different direction than all this deep subtle complex MeTTa-ness -- making it natural to reflect ad comment a bit on how Hyperon fits into the LLM landscape.   "Going beyond the shortcomings of LLMs" is totally not how Hyperon was originated and certainly not the most natural way to think about Hyperon, but it's a meaningful angle to take thought-experimentally, especially in the context of building practical systems integrating LLMs with Hyperon in various ways.

LLMs have an interesting positioning in the context of general intelligence. LLMs can be considered narrow AI in a principled sense, yet they possess an incredibly broad scope in a human sense.  They operate by utilizing vast amounts of training data and exhibit limited generalization beyond that.  The fact that LLMs don't venture far beyond their training data distinguishes them from powerful AGI, which inherently involves generalizing beyond initial programming and training data. However, LLMs, by generalizing just slightly beyond their extensive training data, can perform a very broad range of tasks, because their training data covers so much of what interests humans in their everyday pursuits.

It is also notable and fascinating that LLMs display some forms of emergence. For example, they demonstrate few-shot and in-context learning by adapting to new examples without modifying the weights of the network, purely by dynamics in their neural activation space.  While the concept of learning in a neural network without weight modification is not new, the scale at which LLMs are able to apply this type of learning is unprecedented.

However, there are some quite serious limitations to LLMs. For instance, systematic multi-step reasoning, as required in scientific research or producing groundbreaking mathematical theorems, is challenging for  these systems. This is partly because they predominantly recycle existing knowledge rather than extending far beyond it. Additionally, LLMs struggle with creativity, often producing work that is derivative rather than is aesthetically rich  or moving let alone artistically groundbreaking.  In the end, it's clear the major limitations of LLMs are closely related to their narrowness, their training-data-bound characteristics.

Among the overall AI research community, opinions vary widely regarding the relationship between LLMs and AGI. Gary Marcus and Yann LeCun for instance view LLMs as a diversion from the path to AGI, while others think of them as significant progress, believing that scaling up and augmenting current LLMs, or combining LLMs with other technologies, could be a viable path to human-level AGI.

As should already be clear and will be made even clearer as this document unfolds, LLMs are far from the core of the Hyperon system or its associated theory.  However, our current working hypothesis is that LLMs may be tightly or loosely integrated with Hyperon systems and in this way could come to play a significant role in Hyperon-based general intelligence.

There are also more general lessons to be drawn from LLMs for the development of Hyperon and other non-LLM-centered AGI approaches.   For instance, one key point driven home by large language models (LLMs) is that -- surprise, surprise! -- massive scaling can sometimes dramatically enhance the capabilities of an AI system.  Hyperon, like modern deep neural net frameworks but with subtler particulars, is built with scalability in mind. It leverages modern hardware, distributed processing, and even blockchains to achieve this.  The impressive scalable performance of LLMs provides a valuable pressure to leverage modern computing and data infrastructure technology to bring other AI algorithms to a similar level of scalability, so they can start to fulfill their historical promise as LLMs have allowed deep neural nets to do, and so they can effectively hybridize with LLMs to yield emergent forms of intelligence.

Digging a little deeper, Alexey Potapov summarizes some of the reasons for pursuing Hyperon as an alternative or augmentation to LLMs as follows:

{\tt \small
"Limitations of LLMs are now becoming generally clear, alongside their impressive strengths. Short-term results can be achieved by integrating symbolic knowledge and reasoning with the existing pre-trained LLMs. However, integrating the existing LLMs (especially via prompts) even with an advanced knowledge graph and a symbolic reasoning system is arguably not enough for achieving human-level AGI.
 
"The focus in AGI R\&D can be different. Improvement of LLMs with symbolic components in the context of Hyperon development may include augmenting LLMs with external rewritable memory based on metagraph querying. Alternatively, DNNs, even such large as LLMs, can be considered as specialized modules (a sort of reflexes, instincts, or skills ? general narrow AI rather than AGI or even narrow AGI by themselves), which are controlled by explicit knowledge and by reasoning-based engines integrated with other approaches and techniques. Using Hyperon makes great sense within the latter approach, though there are also sizable challenges.
 
"Any scenario of AGI development will arguably include achieving capabilities of dealing with real-world situations, which are richer than natural language descriptions. The Hyperon-based approach supposes learning explicit representations of such situations, which should be not as brittle as the contemporary symbolic representations and much more structured that implicit representations learned by neural networks. Thus, two main challenges for Hyperon are:

\begin{itemize}
\item   	Scaling knowledge metagraphs, at least, to the amount of information digested by LLMs;
\item  	Mitigating the brittleness sometimes found in symbolic AI methods, which sometimes comes together with their dependence on hand-crafted representations, rules and algorithms.
 \end{itemize}

"Overcoming both challenges might be possible within different approaches and, in particular, with different roles for DNNs/LLMs and different types of neural-symbolic integration; choosing the best approach is a challenge by itself."
}

\subsection{Specialized Spaces with Diverse Roles}

Returning now to our explication of the technical guts of the Hyperon approach to AGI -- The primary meta-representational hub of Hyperon is the Atomspace metagraph, but there is also a broader Space API that permits the creation of multiple specialized types of Spaces within it, complementing the standard in-RAM Atomspace.   These additional Spaces can serve various functions, such as distributed pattern storage and look-up, efficient concurrent execution, and integrating neural networks in a manner that makes them appear as Atoms.

The variety of Spaces is part of what allows Hyperon and  MeTTa to serve not only as theoretical framework and research tool but also as an infrastructure for practical real-world applications. Its flexibility allows it to serve various purposes ranging from creating self-modifying code soups to serving as a smart contract language in a blockchain. With advancements in computer hardware, such as increased RAM and faster processors, the viability of using MeTTa for a plethora of applications becomes even more feasible.

For example, there is a distributed Atomspace (DAS) that can run across multiple machines. This variant of the Atomspace backends on MongoDB and Redis, using modern NoSQL database technology to have a large distributed Atomspace spanning across multiple machines. 

There is also a Rholang Atomspace, which contains MeTTa programs compiled to Rholang, a language known for running various programs efficiently on concurrent processing infrastructures, such as multicore machines, and cores with built-in multithreading capability (and experiments are currently being run with Rholang on the cutting-edge APU Associative-Processing-Unity chip by GSI) . By creating what is termed a "Rholang Atomspace," one can compile MeTTa-programs into Rholang, which is then executed efficiently on a concurrent processing infrastructure. This brings with it the advantages of efficient concurrent processing and integration with blockchain technologies and encryption, among other features.

Another key functionality elegantly enabled via the option for multiple types of Spaces is the capability of interfacing with neural networks through the Atomspace API. This means that you could wrap a large language model or other deep neural networks in an Atomspace API, and perform pattern matching against them. This creates a bridge where there is a translation between formal pattern-matching queries and, for instance, natural language queries or activations in a neural network. This results in the formation of what can be referred to as a neural Atomspace, which serves as a neural lobe within the overall Hyperon system.  A single Hyperon instance could potentially contain multiple neural lobes implementing different neural architectures oriented toward different purposes; the neural lobes could interconnect directly and/or use other spaces such as the default Atomspace as a hub.

There is also an initiative aimed at working with specialized custom hardware to improve efficiency and scalability: the (currently in the early stages of development)  OpenCog pattern-matching chip.  In a collaboration between Simuli and TrueAGI, this chip is being designed to contain an on-chip Atomspace that supports Hyperon-style pattern matching, albeit with potential restrictions on the API based on hardware constraints.

\subsubsection{The Distributed Atomspace}

\begin{figure*}
\begin{centering}
\includegraphics[width=15cm]{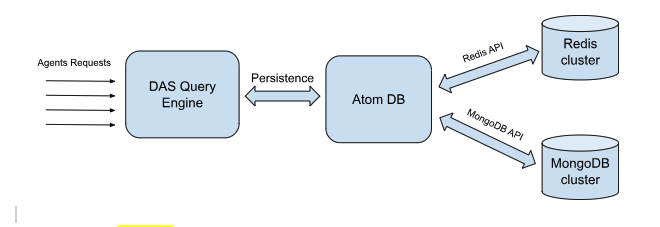}
\protect\caption{\label{fig:DAS} High level architecture of Hyperon Distributed Atomspace}
\end{centering}
\end{figure*}

Andre Luiz de Senna, one of the lead developers of the Webmind Cognition Engine and Novamente Cognition Engine that preceded OpenCog, and of the original version of the OpenCog Atomspace, has more recently led the design and development of Hyperon's Distributed Atomspace (as loosely depicted in Figure \ref{fig:DAS}.    He summarizes his work on this as follows:

{\tt \small "Distributed Atomspace (DAS) is the hypergraph used by OpenCog Hyperon to represent and store knowledge. The nature and the amount of knowledge that needs to be stored vary a lot depending on the domain and the AI algorithms being used to address the problem but the way the AI agents interact with DAS is always by using it as the source of knowledge and the container of any computational result that might be created or achieved during their execution.

"As a data structure, DAS plays a central role in every AI agent. Operations like hypergraph traversing, querying for node's or link's properties, subgraph matching, etc are performed all the time during the execution of any AI agent, and also requests for addition of new elements and changes in properties or in connectivity of nodes, links and subgraphs as the agents achieve intermediate or final results. Therefore, DAS must have a very flexible API with efficient CRUD operations for hypergraph elements and a robust indexing system to allow efficient execution of complex queries involving properties, connectivity, subgraph topology, etc.

"Keeping such a flexible and responsive API is a great challenge when the size of the knowledge base is very large, demanding a persistence back-end. Using a scalable database engine sounds like a straightforward approach to achieve these goals but there are issues that need to be addressed in order to make them suitable to represent and store knowledge bases in the way DAS requires.

"The choice of which database engine to use is the first one. This is an issue because some aspects of the Atomspace are better represented in some types of databases while other aspects are better represented in others. This is also true for the indexes we need for the queries. Each database engine offers specialized index types which would be very useful like hash tables, text indexes with support to regular expressions, geospatial index, b-trees, schema index, etc but no database engine offers all of them.

"In addition to this, AI agents need to drive their attention to the most relevant portion of the knowledge base at a given moment, focusing on some portions of the hypergraph while disregarding others. Therefore, DAS needs a sort of cache hierarchy capable of keeping the most relevant information closer to the agents (local RAM) while the information which is not being actively used at the moment may be away (remotely or in disk). This is an issue because the principle of locality applied to Atomspace knowledge bases don't follow any of the typical types (temporal, spatial etc) intrinsically assumed by database engines to implement caching and load balance policies, therefore these policies tend to underperform badly in OpenCog Hyperon.

"To address the first issue, DAS uses multiple database engines in the persistence back-end with a layer to abstract their APIs and to route requests to the more suitable engine according to the available indexes, as illustrated in Figure X. Some aspects of the knowledge base, (e.g. nodes and links properties) are modeled in a document DB while other aspects (e.g. hypergraph topology) are modeled in a key-value DB. We use MongoDB and Redis in our PoC implementation but there could be other DB engines as well.

"Requests to the persistence layer are routed to one or more DB engines. For instance, requests for the links in the incoming set of a given node (i.e. links that point to a given node) are redirected to Redis while requests for nodes whose name matches a given regular expression are redirected to MongoDB. The choice is made in the Atom DB based on the available indexes in each DB engine. 

"DAS has a local cache running in the MeTTa interpreter. It's an Atomspace with API to create, modify and query for atoms either by traversing the hypergraph or by subgraph pattern matching. This Atomspace is integrated with the interpreter in a way that the calls to this API are transparent to the MeTTa programmer. Caching policy is defined at runtime using subgraph patterns to determine how atoms move from the local cache to the main DAS server and the other way around.

"There's a Proof-of-Concept implementation of DAS (as shown in Figure X), developed as an isolated component and deployed in a cluster of 5 servers (3 for Redis, 1 for MongoDB, and 1 for DAS) tested with a knowledge base of 300M atoms running simple AI algorithms locally. The response time for typical queries was encouraging so now we're aiming at turning it into a real Hyperon component integrated in the MeTTa interpreter. Basically, the plan is:

\begin{itemize}
\item Implement the local cache which will run inside the MeTTa interpreter, with dynamic cache policy rules based on relevant subgraph patterns.
\item Implement a scalable deployment architecture capable of auto-scaling to fit knowledge base size as well as query load.
\item Tune Atom DB and the DB engines in order to make the most of all the indexes and clustering features offered by each engine
\item Extend DAS Query Engine to support all query types required by Hyperon AI agents
\end{itemize}
}

\subsection{Decentralized Deployment via Blockchain Integration}

Through appropriate blockchain integration, a distributed Hyperon instance can be spread across many machines worldwide, without need for any single owner or central controller. One can have coordination among many agents running on different machines without requiring complete trust between the parties involved.   

The use of blockchain infrastructure has a variety of advantages.   It opens up a host of compute resources for usage by distributed Hyperon systems, including spare resources on home computers and phones and dedicated home compute boxes, and server farms historically used for crypto mining.   It also leads naturally to infrastructure that is secure by design, and respects the sovereignty of the individuals and entities who have provided data to train and teach early-stage AI systems.  It encourages the use of creative tokenomic models to incentivize provision of hardware, data and educational interaction to these system.

Ben Goertzel observes that, {\tt \small ''As progress advances further toward human-level AGI, the use of decentralized infrastructure also decreases the odds of any single party achieving autocratic power over powerful AGI systems.   A core intuition here is that decentralized, participant-governed systems like the Internet and the Linux operating system provide a better model for the coordination and operation of powerful AGI than centralized IT systems such as those currently offered by e.g. US and Chinese Big Tech companies.''}

Key decentralized tools currently planned for deep integration into Hyperon architecture include:

\begin{itemize}
\item SingularityNET protocol for coordination among decentralized AI agents
\item NuNet protocol for decentralized coordination of compute resource usage among AI agent populations
\item Hypercycle ledgerless-blockchain protocol for secure scalable decentralized communication among distributed software processes
\item The Rholang language's capability for efficiently executing software programs (''smart contracts'') across decentralized networks 
\end{itemize}

Hyperon itself can operate perfectly well without any of these.   However, it would need to be deployed in a traditional way on centralized server farms, with security provided via firewalls and traditional mechanisms, and with difficulties operating integrated cognitive networks with different portions owned by different individuals or entities.

Rholang has a special role here in that it has two uses for Hyperon: to enable secure decentralized execution of MeTTa scripts, and also to enable efficient execution of MeTTa scripts on concurrent processing infrastructure.

SingularityNET's Chief Product Officer Jan Horlings summarizes the value added by these various tools as follows:

{\tt \small
'' At its core, the SingularityNET platform is a distribution channel for AI services that allows any developer of any sort of AI method to monetize API calls, in a trustless setting. Once published, any user can independently start making API calls and the publisher/developer will receive AGIX based on successful calls made, without additional overhead. On the platform, Hyperon will be a very sophisticated and general AI service among many smaller, more dedicated AI services and Knowledge Graphs. 

"Hyperon will be offered on the platform with a number of AI strategies and functionalities deeply integrated, such as LLM, ECAN, PLN and SISTER. There will however be opportunities to integrate a wider variety of services into the Hyperon Framework. The decentralized platform offers this opportunity to every developer or organization that would like to take part in extending the capabilities of Hyperon. (Or, framed differently, extend their own service by integration with Hyperon)

"A crucial enabler of the platform dynamic between services and Hyperon is what we call ''AI-DSL'' or AI Domain Specific Language. We are developing this sophisticated intelligent orchestrator of services that will recognize the inputs, outputs, and purpose of each service, as well as other attributes, such as a multifaceted ?reputation' score. It will be able to create ad-hoc workflows based on available services and user requests. LLM-based enhancements of AI-DSL will enable AI users to give the platform instructions in natural language; the Platform Assistant. It will figure out what the ?best' available services are (fastest, cheapest, highest quality, etc,) depending on the use case and the user's specific needs) and the best sequence to reach the desired result. 

"In other words, once we have a sufficiently mature version of Hyperon running on the platform, the Platform Assistant (AI-DSL) will function as a User Interface to Hyperon and other dedicated services on the platform. 
Of course, this is not the only way to run dedicated services or Hyperon on the platform. All kinds of tools and applications can interface with hyperon or a specific service directly or can rely on an API-based version of AI-DSL to continuously monitor the platform AI-service ecosystem and come up with the best sequence at a given time. 

"This way the platform firmly embeds Hyperon in an open and fully decentralized ecosystem of -potentially very diverse- Knowledge Graphs and AI services, enabling the global community of developers to contribute and enhance its core capabilities and similarly, enabling anyone to connect and benefit from its growing potential. 
}

The basic strategy for integrating Hyperon with these decentralization-oriented tools is relatively simple.   Each DAS component, and each local Atomspace (with coupled MeTTa interpreter), can be treated as a SingularityNET agent, wrapped in a Hypercycle AI Machine container, which is associated with a NuNet node that manages its deployment and orchestration.   The MeTTa scripts running in one of these Hyperon agents, when they need to interact with other agents in the network, will be compiled to Rholang and use Rholang's integration with HyperCycle to carry out secure messaging with other agents.  

\begin{figure*}
\begin{centering}
\includegraphics[width=15cm]{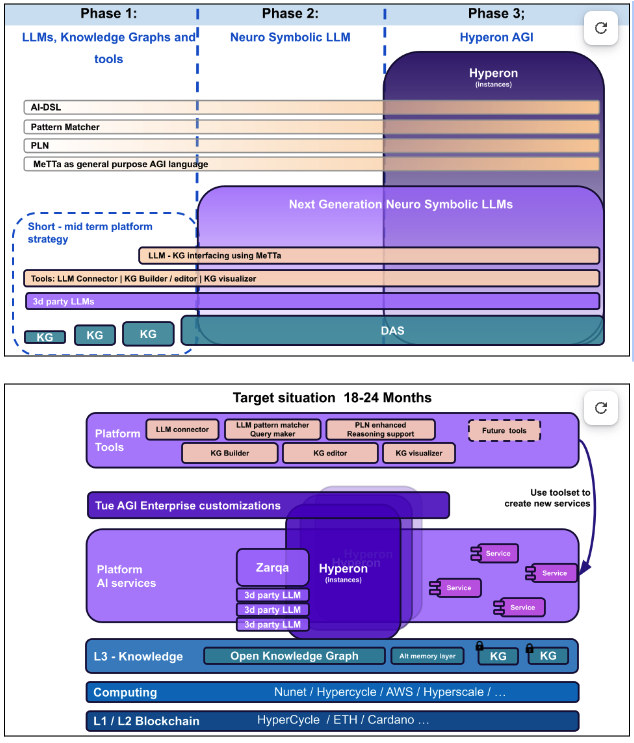}
\protect\caption{\label{fig:SNet-Roadmap} Rough tentative roadmap for some of the key development initiatives regarding the SingularityNET platform}
\end{centering}
\end{figure*}

Implementation of all this, of course, involves many non-simple aspects and will be carried out gradually as the various tools involved move from alpha to beta to full production versions over 2023-25.   In parallel with the maturation of Hyperon, the various SingularityNET ecosystem infrastructure tools are being rapidly built out.  As roughly indicated i Figure \ref{fig:SNet-Roadmap}, the SingularityNET platform strategy for 2024 involves a particular focus on decentralized hosting and utilization (within LLMs, early Hyperon versions and other AI tools) of knowledge graphs covering various particular domains, segueing as Hyperon matures to a focus on decentralized hosting and interconnection of AI services playing a ''plugin'' role to decentralized Hyperon instances.   The various vertical market specific projects spun off from SingularityNET Foundation (e.g. SingularityDAO in the DeFi domain, Rejuve Network in the longevity domain, Mindplex in the media arena) will then fulfill a portion of their business models via offering appropriate Hyperon plugin agents on SingularityNET / Hyperon / NuNet infrastructure.

\subsection{Compiling MeTTa to Rholang for Rapid Secure Decentralized Execution}

An additional twist to the infrastructure and formalism of Hyperon is the ongoing project to compile MeTTa into the Rholang language, which is aimed at two purposes:

\begin{itemize}
\item enabling efficient concurrent execution of MeTTA programs on appropriate hardware
\item enabling MeTTa's use as a smart contract language for the HyperCycle ledgerless blockchain (and potentially other chains as well)
\end{itemize} 

In some aspects the latter application is orthogonal to MeTTa's use as an AGI language -- it was noted above that the unique advantages and aspects of MeTTa as a programming language have some broader applicability beyond the AGI sphere.  For instance, using MeTTa to write smart contracts for ledgerless DeFi makes total sense independently of AGI or even AI -- because MeTTa lends itself to formal verifiability and efficient execution, both of which are critical in a DeFi domain.  

However, there are also a lot of obvious potentials for synergy in the dual use of MeTTa for AI scripting and AI-generated code within Hyperon systems, and for scripting operations of HyperCycle networks in which Hyperon systems are embedded.   There is significant potential here for developing distributed AI systems with significant amounts of emergent intelligence at the decentralized-network level.

The MeTTa/ Rholang interfacing project is being carried out in collaboration between SingularityNET, Hyperon and Lucius Greg Meredith and his company F1REFL3Y, which is the successor to his previous project RChain.   RChain developed a novel blockchain technology centered on the Rholang smart contract language, a syntactic front end for Meredith's rho calculus, a novel mathematical calculus breaking new ground in the formalization of concurrent computational processes.   

Rholang demonstrates remarkable properties in the area of secure, efficient execution on concurrent hardware, including in a decentralized context.   During 2023 Meredith and his team have been developing a source-to-source compiler from MeTTa to Rholang, with the objective of granting compiled MeTTa the same concurrency and security advantages of Rholang.

\begin{figure*}
\begin{centering}
\includegraphics[width=15cm]{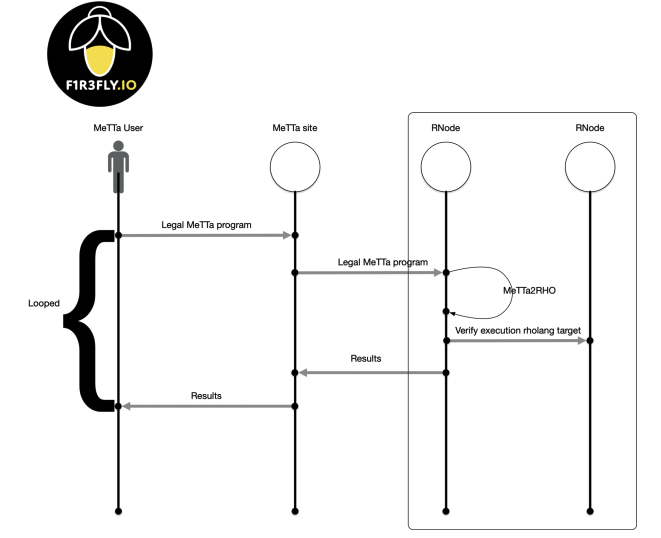}
\protect\caption{\label{fig:RNode} High level depiction of the control flow relating MeTTa to Rholang in the Hyperon architecture.}
\end{centering}
\end{figure*}

\begin{figure*}
\begin{centering}
\includegraphics[width=15cm]{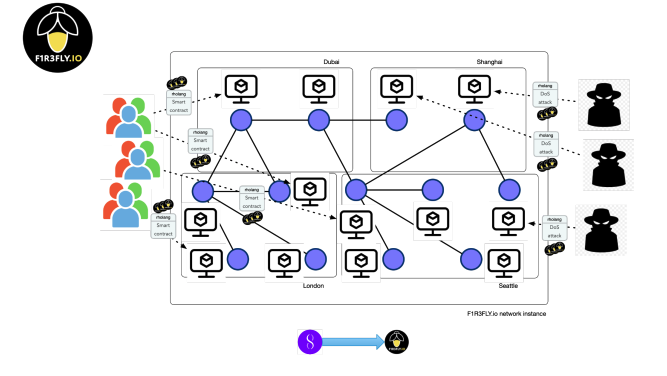}
\protect\caption{\label{fig:RNetwork} High level depiction of a network of RNodes, used for evaluating distributed Rholang programs in a secure way.   These. RNodes may interact according to a number of possible blockchain infrastructures, including Hypercycle.}
\end{centering}
\end{figure*}

Meredith summarizes some key aspects of this work as follows:

{\tt \small
"If AI is going to be decentralized, it has to take on a lot of the characteristics of the blockchain.
In particular, a node that runs some AI resource, like an LLM, or a theorem prover, or some
combination of AI resources, will need to be robust against denial of service attacks. Likewise, a
network of such resources will need to be robust against the failure of any one node in the
network. This is one of the motivations for integrating SingularityNet's MeTTa language into
F1R3FLY.io's RNode.

"The other reason is performance. At current performance RNode gets about 1000 transactions
per second (tps) per node per processor, where a transaction is measured as a committed and
durable communication of data from a data provider to a data consumer. This execution model
is implemented in terms of a novel key-value database (KVDB) called RSpace. The main
novelty is that both data and continuations are stored at keys in RSpace. The execution
mechanism is programmed via a smart contracting programming language called rholang.
The MeTTa2Rho compiler compiles the language MeTTa into rholang. As a result, MeTTa
execution scales as a network adds nodes that also add processors, because rholang execution
scales as the network adds nodes that also add processors. However, the benefit of compiling
MeTTa into rholang doesn't just stop at scaling.

"Indeed, the network of RNodes are all running under a consensus algorithm called
CBC-Casper. This ensures that all the nodes have copies of the same execution state and as a
result, if one or nodes fail, the network can continue to operate. Further, to execute some
rholang a node requires the client to pay for the computation and storage associated with a
token associated with the network. This constitutes a prophylactic against inevitable denial of
service attacks."}

Figure \ref{fig:RNode} gives an overview of the control flow involved here.   As Meredith explains, {\tt \small "Per the diagram a MeTTa client (which could be a human user or a computational agent)
submits a legal MeTTa program to the system. In this version we assume it first hits a gateway
which forwards it on to a node in the network. However, it is perfectly legitimate to have the user
submit the program directly to a node. These nodes are modified versions of RNode that
contain a MeTTa2RHO compiler. The compiler transforms the MeTTa program into rholang
which is then run against the node.

"In the production implementation the execution produces a run log which is then verified by all
the other nodes in the network. If there are any results, they can be safely communicated back
to the user at the end of the verification phase, or optimistically communicated at the end of the
target node's execution of the rholang." }

Figure \ref{fig:RNetwork}  loosely depicts the network of RNodes.  As Meredith notes, {\tt \small A network for RNodes (often called a shard) consists of some deployment of nodes in various
geographic locations, e.g. some nodes in AWS US-West, some nodes in GoogleCloud UK,
some nodes in IBM Cloud EU, etc. These nodes are linked by an implementation of Kademlia,
together with CBC-Casper. A client signs a request for execution. The signature is used to find a
wallet that must contain enough tokens to execute the compute and storage associated with the
rholang."

"The compiler being developed from MeTTa to Rholang is detailed in the paper{\it Meta-MeTTa: An operational semantics for MeTTa} \cite{meredith2023metametta}.   While the details are quite involved the core approach is relatively simple:
Essentially, MeTTa's semantics is given by a register machine; and that register machine is easily
compiled into rholang in a correct-by-construction design pattern.}

The process of making this compilation work has also been valuable in terms of shaping the feature set of MeTTa itself; for instance 

\begin{itemize}
\item Sealed terms: To prevent unwanted information leakage future versions of MeTTa will support a means of
sealing a term from outside probes.
\item  Comprehensions: To allow for defensible communication with resources across a trust boundary future versions of
MeTTa will support a notion of comprehensions.
\end{itemize}

\noindent have both been added into MeTTa as a result of the F1REFL3Y collaboration, and both promise to have valuable applications beyond the smart-contract application (e.g. as one example, comprehensions provide a simple explicit representation of PLN semantics at the MeTTa level).

\subsection{The Need for a Cognitive AGI R\&D Platform}

A number of difficult, complex choices were faced in choosing the architecture we've described for the Hyperon platform, and we certainly can't say we believe the current Hyperon system is the only workable approach.  However we do feel it has significant advantages to anything else out there today, or anything else whose development we know to be underway.   Alexey Potapov emphasizes the need for a cognitive AGI R\&D platform, at the current stage of evolution of the AI field, via first summarizing some relevant limitations of LLMs as follows:

{\tt \small
"Deep learning has been showing breakthrough results in different domains during the last decade. The recent results achieved by LLMs make us wonder if they bring us close to AGI. However, their limitations, such as lack of memory and world models, are also obvious, and they are increasingly being overcome with the help of symbolic methods through neurosymbolic integration.
 
"There are dozens of wrappers over ChatGPT or LLaMa such as Langchain, AutoGPT, Voyager, etc. They are pieces of ad hoc code in imperative programming languages, which implement different non-composable ways of controlling LLMs. This is needed, because LLMs are a sort of ''language reflex'', which benefits from symbolic control even for specific practical applications.
 
"Let us consider LLMs as a candidate for AGI. Imagine that we want an LLM to play chess. How will it do this? At best, it will generate an external call to Mu Zero. But will it be able to control Mu Zero at each move? No. The LLM will have no idea of the current situation on the game board. Imagine, we ask LLM not just to play chess, but to checkmate with the rook. How will LLM be able to pass this information to Mu Zero? Will Mu Zero be ever able to take this information into account by itself? Not at all, without modifications or possibly retraining. Now imagine that we ask LLM to play on two boards simultaneously and to use the same type of piece during the same turn on two boards. Should the LLM coordinate two instances of Mu Zero on each turn? Yes, but it cannot do this, it has no idea regarding the game situation and internal processes in Mu Zero. In order for an AGI system to do such sort of things, it needs to have a shared representation of different aspects of the world including language, and an integrative decision-making or action-selection process based on it.
 
"An adept of deep learning would say that we just need to train a huge DNN on all the real-world data we have, which will learn such this representation together with strategies of using it by itself. However, such the approach (if it ever workable) is far beyond currently available computational resources and data, while capabilities of modern symbolic systems as components of neurosymbolic cognitive architectures are still highly underutilized and underexplored.
"
}

LLMs are already being used as crude hubs for integrating a diversity of AI methods from different paradigms, implemented as LLM plugins.   However, black-box-style shallow integration via prompts is quite limited in nature, and a cognitively richer architecture can allow combination of methods from different AI paradigms in subtler and more powerful ways.   As Potapov points out,

{\tt \small
"Different paradigms and methods in AI did not arise by chance. They have different strengths and weaknesses. LLMs also quickly became integrated with different external tools, because solving problems in different domains is better done by different means. 
 
"Hyperon supposes a full interoperability of different paradigms. The core operation in Hyperon is retrieving information from  storage in a general way such that this operation covers queries to (meta)graph knowledge bases, pattern matching in functional programming, unification in logical reasoning, and even operations performed by attention heads in Transformer networks or other neural modules. Chaining of such operations allows constructing a Turing-complete language, which serves as a ''cognitive assembler'' and a generic approach to implementing declarative and procedural, episodic and semantic memory together with different forms of processing information in them. Different types of storages can be used simultaneously, and queries to them can be composed. By default, it is a metagraph knowledge base with symbolic queries, Atomspace, which can contain subsymbolic atoms as well. However, entirely neural spaces can also be used based on hyperdimensional vectors, graph embeddings, attention mechanisms or just prompt-based querying. Query chaining is described by the content of spaces (or memory from the cognitive perspective) themselves, which allows them to be implemented in an arbitrary way as well as to be learned. As a result, different strategies of declarative reasoning as well as other inference strategies such as probabilistic or genetic programming or neural module networks can be implemented and combined. For example, neural module networks can be assembled on the fly by symbolic reasoning or by themselves, as well as declarative reasoning can have symbolic or neural implementation.
 
"While we suppose implementing some concrete ideas and AGI theories as Hyperon modules, AGI is still a wide R\&D domain, and Hyperon is a platform precisely for AGI R\&D with the focus on cognitive synergy between different paradigms, approaches, techniques. This makes Hyperon different from other proto-AGI systems, which are typically built around one particular theory and are difficult to use in AGI R\&D, which goes beyond this theory.
 
"While DNN-centric solutions typically use quite weak symbolic overlays, advanced symbolic systems are usually integrated with DNNs in a shallow way. Hyperon allows incremental development of multi-paradigmatic interoperable components with strong both symbolic and neural parts, which makes it not only a platform for research and prototyping, but also a promising framework for developing a concrete AGI system. If s multi-paradigmatic integrative approach is more efficient than mono-paradigmatic, then Hyperon has chances to become a leading AGI platform."
}

\section{The CogPrime Cognitive Model (and Beyond)}

The Hyperon software framework can be adapted to implement a variety of cognitive architectures and AGI approaches.  For instance, it can be used to create chat systems that answer questions in a one-off basis, theorem provers, or even cognitive architectures with resemblance to the human mind.   That said, there is a fairly particular cognitive architecture that has been at the center of OpenCog development since the beginning, continuing into the Hyperon era.   In this section we will discuss this "historical default Hyperon cognitive architecture" and some additional ideas as well.

The most thorough presentation of the recent incarnations of the CogPrime cognitive model are given in Goertzel's 2021 {\it General Theory of General Intelligence} paper \cite{GTGI}, which contains detailed pointers into a number of other relevant recent technical research papers.   The review we're about to give here is less mathematical and higher-level, and aimed more at getting across the basic concepts than giving a rich understanding of the underlying theory.

\subsection{CogPrime: Hyperon's Historical Default Cognitive Model}

CogPrime is a somewhat flexibly defined cognitive architecture that has been centrally pursued throughout the history of the OpenCog project; it was put together in 2012 in the context of OpenCog Classic, but still applies perfectly well in the Hyperon context.  This architecture has been called ''CogPrime'' in various publications, although this name never particularly caught on.  Nevertheless we believe it is important to distinguish OpenCog as a software framework from particular cognitive architectures that may be implemented with in it.   OpenCog Hyperon as a framework has been designed in substantial part to meet the needs of the CogPrime architecture, but there have also been other desiderata in mind, and it is quite possible that Hyperon will be used to explore other architectures besides CogPrime, and/or that CogPrime will heavily evolve as hands-on AGI R\&D with CogPrime in Hyperon proceeds.

The primary reference on CogPrime is the 2-volume book {\it Engineering General Intelligence} from 2014 \cite{EGI1} \cite{EGI2}.  To simplify a lot of complexity, what is going on in CogPrime is:

\begin{itemize}
\item Perception forming Atoms in Atomspace, linked in with representations in other spaces such as neural ones
\item Actions comprised as Atoms in Atomspace, then in some cases translated into other representations for actuation (e.g. neural activation patterns, Rholang programs...)
\item Ambient cognitive activity occurring in Atomspace, e.g. importance spreading (in which various importance values spread among Atoms according to their connections), concept formation (which builds new nodes representing new ideas from existing ones), reasoning (which builds new relationships from existing ones).  This is spontaneous, self-organizing activity, not directly driven in a goal-oriented way.
\item Goal-oriented activity, wherein the system using reasoning to select (and synthesize) actions that it believes have high odds of achieving its goals given its perception of the current context
\end{itemize}

Multiple cognitive processes, represented by multiple MeTTa scripts, combine to make all this happen.   The precise mix of cognitive processes is subject to ongoing experimentation, but an initial very detailed hypothesis is given in {\it Engineering General Intelligenc} from 2014 and summarized as well as more concisely online in \ref{GoertzelOCPWiki}.   A quick summary of some highlights is:

\begin{itemize}
\item ECAN, Economic Attention Allocation, for spreading short and long term importance values among Atoms
\item PLN, Probabilistic Logic Networks, for drawing uncertain-logical conclusions from collections of observations, and from knowledge obtained via natural language, mathematics or other sources
\item Concept blending, ''map formation'' , Occam's-Razor-driven ''concept predicatization'' and other heuristics for forming new concepts based on existing evidence and existing concepts
\item Evolutionary learning for evolving new sub-networks of the Atomspace satisfying given criteria
\item Probabilistic procedure and predicate synthesis, for creating new content based on the probabilistic distributions implicit in the Atomspace and other spaces
\item Pattern mining, for creating new predicates representing observed patterns in Atomspace and other spaces
\item Goal refinement, for creating, eliminating and merging subgoals based on the system's given top-level goals
\item Goal-driven action selection, for choosing actions that appear likely to achieve system goals given the perceived context
\item ''Autopoietic'' systems of rewrite rules that rewrite one another, thus creating autocatalytic systems of intercreating rules (an approach sometimes called ''Cogistry'' in the OpenCog context \cite{goertzel2022viable}, and in some ways resembling the use of Replicode in the Aera cognitive architecture \cite{thorisson2012cognitive})
\end{itemize}

\noindent These are all processes operating within MeTTa-based Atomspaces, which may then interoperate with neural spaces and other resources.

\begin{figure*}
\begin{centering}
\includegraphics[width=12cm]{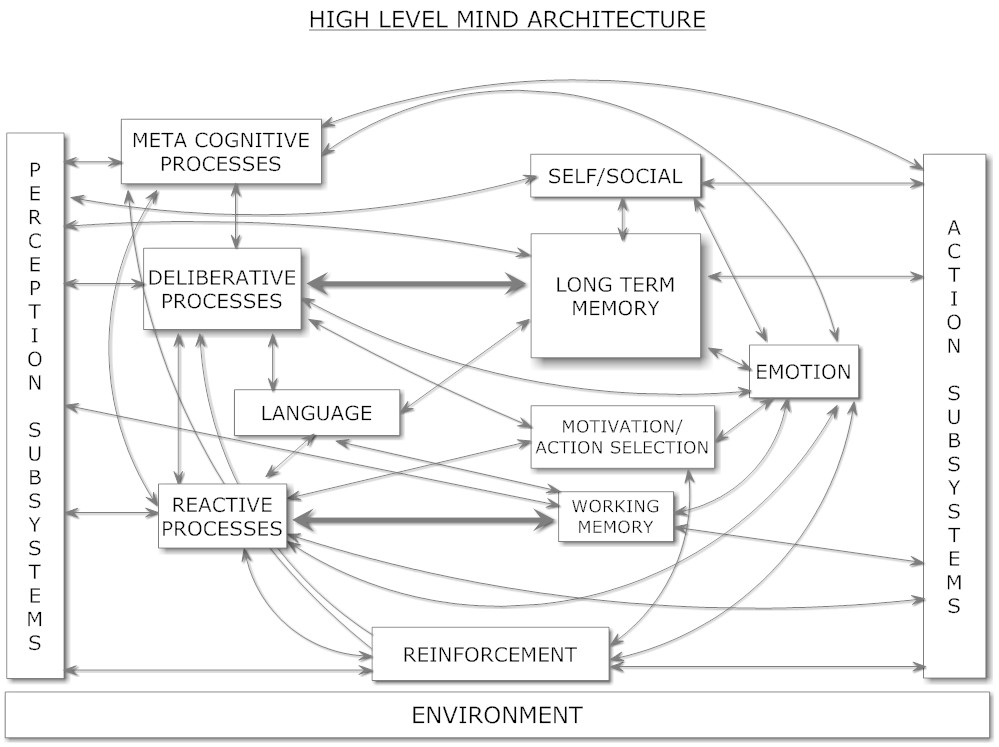}
\protect\caption{\label{fig:mind_arch_sloman} Illustration from {\it Engineering General Intelligence} \cite{EGI2} depicting the key high level components needed to support CogPrime or other similar human-like cognitive architectures.   This is very similar to the basic outline of the Standard Model of Mind, basically filling in a modest amount of additional detail to Figure \ref{fig:sm}.  \newline \newline In the Hyperon architecture, every one of these processes occurs centrally in the Atomspace, initially via a combination of hand-coded MeTTa and then learned MeTTa code filling in the details.   The potential role played by LLMs here is not localized but rather spread among various components --for instance, LLMs may be a big part of the story regarding language; they can be one among multiple forms of long-term memory; they can serve as a source of both reactive and deliberative processes, though with limitations in both cases that indicate they should be coupled with other reasoning and learning processes.}
\end{centering}
\end{figure*}

\begin{figure*}
\begin{centering}
\includegraphics[width=12cm]{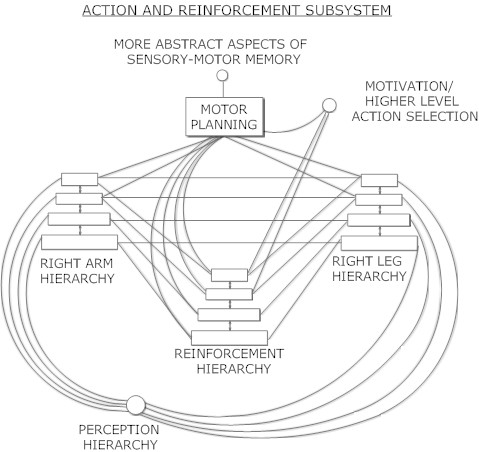}
\caption{\label{fig:mind_arch_action} Illustration from {\it Engineering General Intelligence} \cite{EGI2} depicting the key components involved in carrying out action processes within CogPrime or other similar human-like cognitive architectures. \newline \newline One way to implement this using current technology is to leverage deep neural networks for the action (e.g. right and left arm) and reinforcement hierarchies, and to use Atomspace based MeTTa procedures as the core modality for motor planning.  The magic that needs to happen here is fluid coordination between the symbolic action plans in MeTTa and the neural nets' capability for movement synthesis.   It can't just be a matter of: MeTTa emits a high level plan, and then the neural net figures out how to separately carry out each action in the high level plan.  Rather, the high level plan as a whole has to be taken by the neural nets as context for the synthesis of actions corresponding to each part of the plan, which can then allow detailed motor execution in a way where each sub-action is carried out in a way that reflects the overall movement series of which it is a part.  \newline \newline  This is a fantastic use-case in which to explore neural-symbolic integration, and one we plan to explore in a physical robotics context via collaborations with Hanson Robotics,  Awakening Health, Mind Children and other partner projects; it will be desirable for the open robotics community to become involved in customizing Hyperon for such purposes as well.  }
\end{centering}
\end{figure*}

\begin{figure*}
\begin{centering}
\includegraphics[width=12cm]{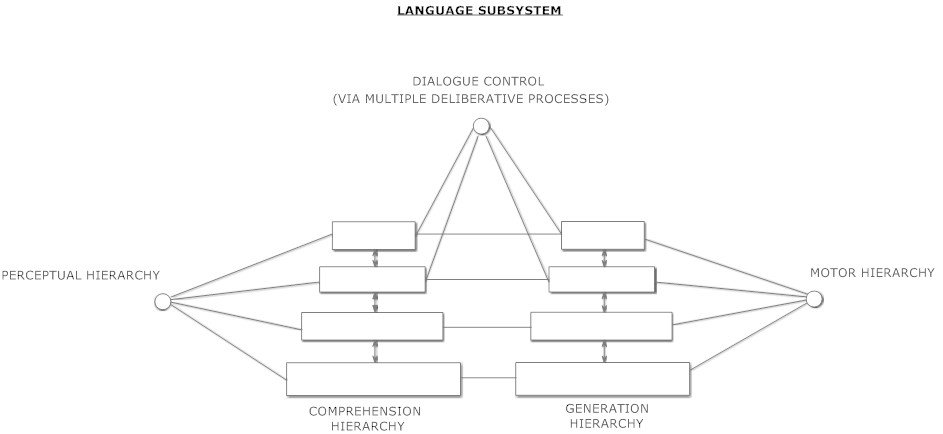}
\protect\caption{\label{fig:mind_arch_language} Illustration from {\it Engineering General Intelligence} \cite{EGI2} depicting the key components involved in carrying out language processes within CogPrime or other similar human-like cognitive architectures.  \newline \newline Transformer neural networks such as LLMs obviously comprise one highly effective way of implementing coupled comprehension and generation hierarchies (where the two hierarchies are richly interpenetrated. more than separated and then closely linked).   If symbolic pattern recognition and inference are used to recognize (probabilistic and/or crisp) formal linguistic rules corresponding to the content within an LLM, then  these inferred rules will also presumably be naturally arranged in a hierarchy, most likely with the same rules predominantly used on both the comprehension and generation sides. \newline \newline Close linkage with perception, action and cognition is clearly critical to language and is under-emphasized in current LLMs.   Part of the cure for LLM hallucination is surely inferential connection of LLMs with knowledge graphs containing assumed ground truth, but connection of LLMs with direct perceptual and active groundings can be another valuable way to supply the needed connection between LLM patterns and non-linguistic realities.}
\end{centering}.
\end{figure*}

\begin{figure*}
\begin{centering}
\includegraphics[width=12cm]{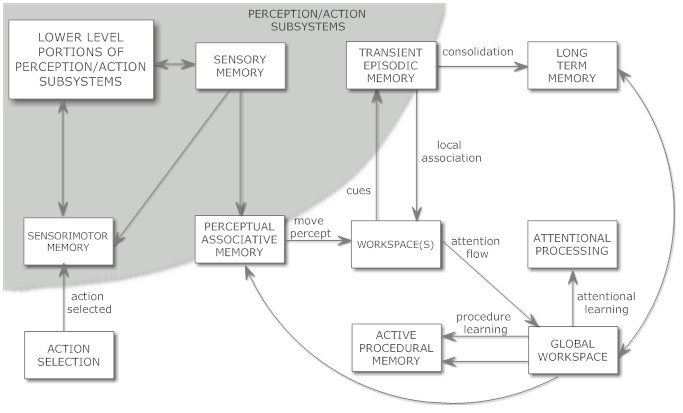}
\protect\caption{\label{fig:mind_arch_LIDA} Illustration from {\it Engineering General Intelligence} \cite{EGI2} depicting the key components involved in working-memory processes within CogPrime or other similar human-like cognitive architectures. \newline \newline In Hyperon the majority of these components are implemented within Atomspace, and their presence in ''working memory'' is indicated by their possession of Short=Term Importance values above a critical threshold (the Attentional Focus Boundary).   Associative memory may be efficiently implemented via hypervector embeddings of Atomspace.  Aspects of sensory, sensorimotor, action and linguistic memory may be stored in neural networks or other subsymbolic components; however it's important that aspects of these are also represented in symbolic form so they can be flexibly manipulated. \newline \newline The dynamics of the Global Workspace Theory \cite{Baars2009}, which are diagrammatically suggested here, are manifested in Hyperon most proximally via the spreading of importance values between the Atoms in the Attentional Focus and those outside it, according to the ECAN (Economic Attention Allocation) equations.\newline \newline LLMs lack a working memory in any richly structured sense, which is part of the reason interacting with them feels more like interacting with a utility than with another human-like cognitive agent.   Various projects building interactive characters based on LLMs are building various sorts of external working memories to cooperate with LLM dynamics.   However, in the end a working memory can't really do its job without a fairly flexible symbolic representation, because part of what has to happen inside a working memory is that the different items it contains need to get varied and combined in a diversity of ways (and efficient flexibility of manipulation is almost equivalent to "symbol-ness")  }
\end{centering}
\end{figure*}

\begin{figure*}
\begin{centering}
\includegraphics[width=12cm]{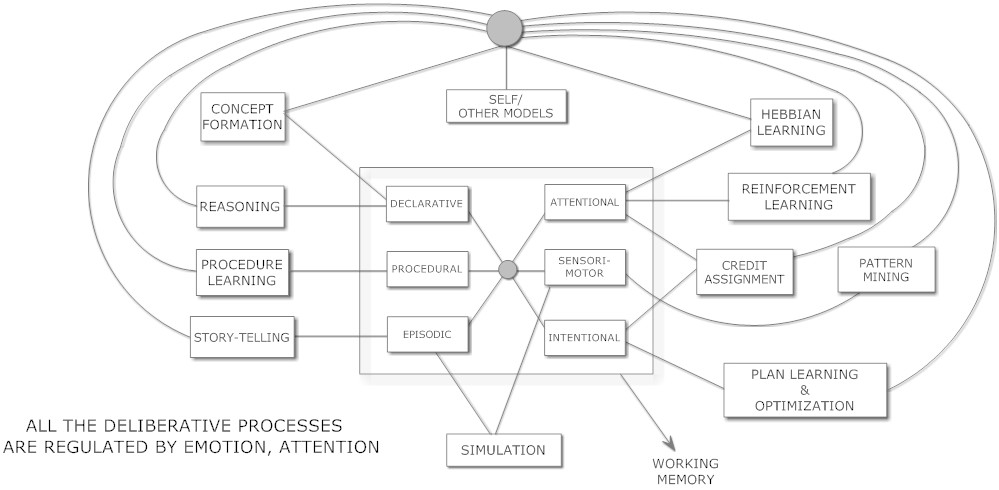}
\protect\caption{\label{fig:mind_arch_LTM} Illustration from {\it Engineering General Intelligence} \cite{EGI2} depicting the key components involved in carrying out cognitive processes associated with long-term memory within CogPrime or other similar human-like cognitive architectures. \newline \newline In Hyperon, this is in many ways the crux of what Atomspace does -- it serves as a long-term memory including knowledge of multiple different kinds, handling them all according to a shared meta-representational fabric (of typed metagraph nodes and links); and it serves as a store as well for the procedures enacting cognitive processes such as the ones filling the roles articulated in this diagram.   \newline \newline In many cases the same Hyperon algorithm can fulfill wholly or partially many of the functions indicated here.  For instance PLN can help with reasoning, procedure learning, storytelling, reinforcement learning, credit assignment and planning.   Evolutionary learning can help with procedure learning, reinforcement learning and concept formation.   Etc.  \newline \newline Many of the functions indicated here are carried out in Hyperon by multiple concurrent and/or cooperating processes -- e.g. concept formation may happen by evolutionary learning, by concept blending, by (e.g. paraconsistent uncertain) formal concept analysis, or by a variety of other heuristics.  \newline \newline For human-like cognition to happen, we need all these processes occurring concurrently in the same large Atomspace metagraph -- this is how you get the cognitive synergy that leads to emergence of large-scale mind-structure patterns such as a self-model or an active self-modifying concept hierarchy/heterarchy.  }
\end{centering}
\end{figure*}

\begin{figure*}
\begin{centering}
\includegraphics[width=12cm]{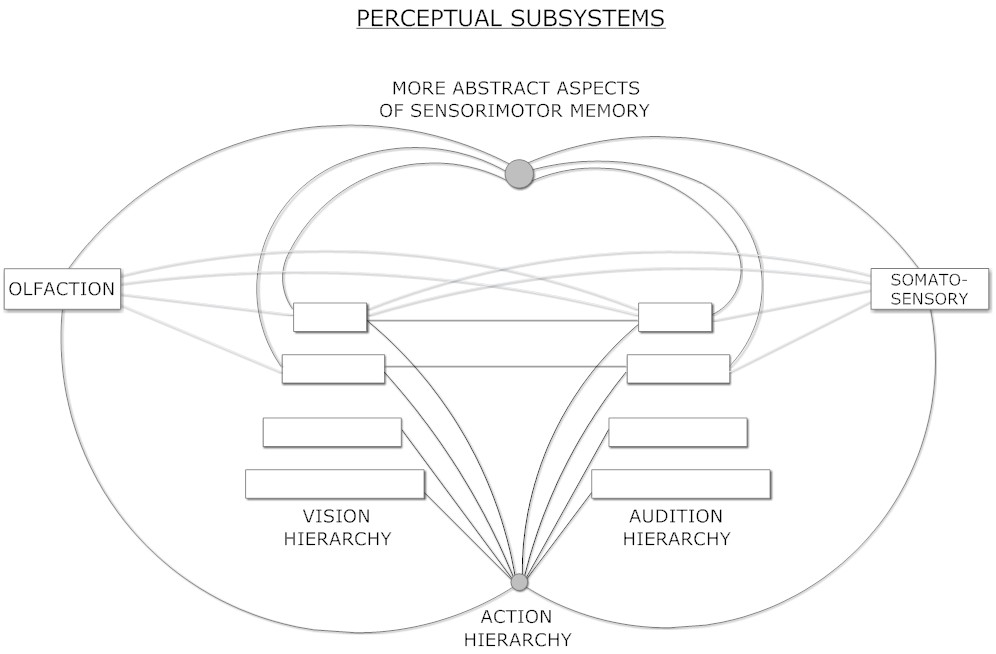}
\protect\caption{\label{fig:mind_arch_perception} Illustration from {\it Engineering General Intelligence} \cite{EGI2} depicting the key components involved in carrying out perception processes within CogPrime or other similar human-like cognitive architectures. \newline \newline In the modern AI world, visual and auditory perception are in many ways effectively handled by hierarchical neural networks trained on large corpora.   However, we believe there are also levels of perceptual understanding that can more effectively be achieved via linking these hierarchies in with symbol hierarchies in Atomspace, which explicitly represent the compositional structure of sensory data. \newline \newline Olfactory and somatosensory perception are less hierarchical in nature; indeed there is some evidence that olfactory pattern recognition in human cortex is based on nonlinear dynamics and strange attractor  or transient formation \cite{Freeman1995}.   Clearly neural models of these are possible but are not currently existent in mature form.   Symbolic understanding may be more valuable in terms of interconnecting lower-level sense perceptions among these very differently organized modalities; and this may be even more strongly the case for non-human AGIs which may have a much greater diversity of sensory channels.}
\end{centering}
\end{figure*}

Figures \ref{fig:mind_arch_sloman}, \ref{fig:mind_arch_action}   , \ref{fig:mind_arch_language}  , \ref{fig:mind_arch_LIDA}   , \ref{fig:mind_arch_perception}, \ref{fig:mind_arch_LTM}  summarize the key components involved in carrying out human-level general intelligence functions according to the CogPrime cognitive model.  The relationships between these cognitive components and software components gets a bit complex -- some correspond to specific software processes, but many are intended to be contributed to by multiple OpenCog software processes, and in many cases the same OpenCog software processes may underlie multiple of these functions.   For instance the Hyperon Atomspace (a software component) is used for both declarative and procedural memory (cognitive components).   Natural language comprehension may be carried out by a combination of LLMs with Atomspace-native processes like PLN.   Etc.   The CogPrime AGI design as presented in {\it Engineering General Intelligence} comprises a detailed theory of how to achieve these various cognitive functions using a. particular set of AI processes, centered on the ones in the bullet list just above.

\begin{figure*}
\begin{centering}
\includegraphics[width=12cm]{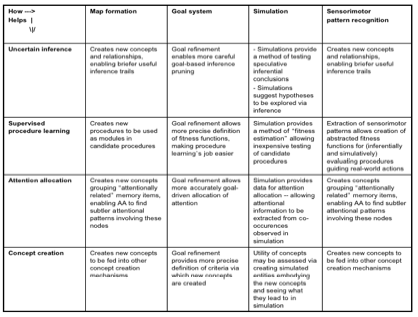}
\protect\caption{\label{fig:cogprimesyn1} Illustration from {\it Engineering General Intelligence} \cite{EGI2} depicting the synergetic dependencies between some of CogPrime's cognitive processes.}
\end{centering}
\end{figure*}

\begin{figure*}
\begin{centering}
\includegraphics[width=12cm]{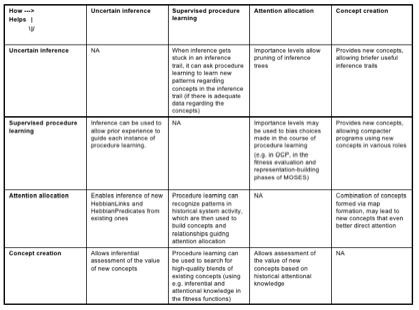}
\protect\caption{\label{fig:cogprimesyn2} Illustration from {\it Engineering General Intelligence} \cite{EGI2} depicting the synergetic dependencies between some of CogPrime's cognitive processes.}
\end{centering}
\end{figure*}

\begin{figure*}
\begin{centering}
\includegraphics[width=12cm]{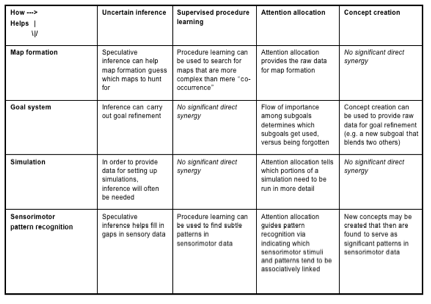}
\protect\caption{\label{fig:cogprimesyn3} Illustration from {\it Engineering General Intelligence} \cite{EGI2} depicting the synergetic dependencies between some of CogPrime's cognitive processes.}
\end{centering}
\end{figure*}

\begin{figure*}
\begin{centering}
\includegraphics[width=12cm]{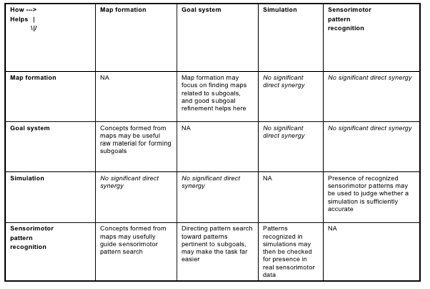}
\protect\caption{\label{fig:cogprimesyn4} Illustration from {\it Engineering General Intelligence} \cite{EGI2} depicting the synergetic dependencies between some of CogPrime's cognitive processes.}
\end{centering}
\end{figure*}

It's convenient to present cognitive architectures in terms of box-and-line diagrams, but of course much of the magic of cognition happens in the interactions and interdependencies between the contents of the various boxes.  Along these lines, a key concept underlying the CogPrime design is that of "cognitive synergy", which has been formalized using enriched categories \cite{DBLP:journals/corr/Goertzel17} but intuitively is very simple to understand: It just means that the various cognitive processes work together rather than against each other, so that e.g. when one of them gets stuck in carrying something out, it can translate its intermediate state into the native languages of other cognitive processes and ask them for help.   Figures \ref{fig:cogprimesyn1}, \ref{fig:cogprimesyn2}, \ref{fig:cogprimesyn3}, \ref{fig:cogprimesyn4}, drawn from {\it Engineering General Intelligence}, illustrate some of the synergies between specific cognitive processes that are suspected to be critical for achieving advanced AGI using the CogPrime approach.

\subsection{Toward a General Theory of General Intelligence}

Goertzel's 2021 {\it General Theory of General Intelligence} \cite{GTGI} paper, building on a number of just-prior research articles \cite{goertzel2021patterns} \cite{goertzel2021paraconsistent} \cite{Goertzel2020metagraph} \cite{goertzel2020grounding}, endeavors to formulate the core notions of the CogPrime architecture within a common elegant mathematical framework, aiming both to elucidate the key ideas underlying the diverse memory, learning and reasoning mechanisms involved, and to make it more straightforward to create highly efficient and scalable implementations.

The GTGI approach involves more precise formulations of the notion that different types of memory relevant to human-like cognition can be represented as different type systems within a typed metagraph, and morphisms then articulated between categories corresponding to these different type systems.   It is also argued that, to a reasonable degree of approximation, the core algorithms required for human-like cognition related to the various needed memory types can be represented by way of Galois connections as various sorts of folding and unfolding operations on metagraphs.   This suggests that seemingly diverse cognitive processes like hierarchical perception and action learning, logical reasoning and evolutionary program learning can perhaps all be implemented efficiently if one has an efficient substrate for carrying out the appropriate forms of folding and unfolding on large metagraphs.   This observation and other more sophisticated variations played a large role in the basic design of the MeTTa language, which provides an infrastructure in which these sorts of metagraph operations can be implemented in a way that is both suitably abstract and appropriately scalable.

\subsection{Hyperon, CogPrime and the Standard Model of (Human-Like) Mind}

As we've noted above, the Hyperon/CogPrime approach to cognition arose via a combination of sources, e.g. philosophy of mind, cognitive science, computer science and mathematics, linguistics, and so forth.   The underlying reasoning is too rich and diverse to concisely summarize here.   However,  to keep things relatively simple for expository purposes, one useful way to consider this cognitive activity is by comparison to what's known about the human mind (i.e. to focus on the a cognitive science angle).   

A thorough analysis of CogPrime in the context of all the processes depicted in Figures \ref{fig:mind_arch_sloman}, \ref{fig:mind_arch_action}   , \ref{fig:mind_arch_language}  , \ref{fig:mind_arch_LIDA}   , \ref{fig:mind_arch_perception}, \ref{fig:mind_arch_LTM}   and their principal subprocesses would go far beyond the scope of an overview like this; basically that is the topic of {\it Engineering General Intelligence, vol. 2}.   A more concise if less informative approach is to look at a simplified model of human-like cognition such as the so-called ''Standard Model of Mind''.

\begin{figure*}
\begin{centering}
\includegraphics[width=12cm]{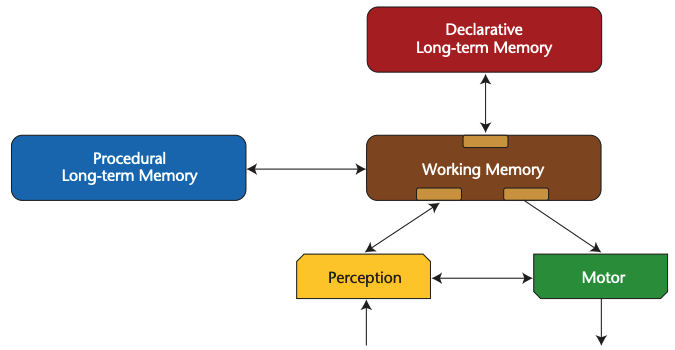}
\protect\caption{\label{fig:sm} Standard Model of Mind: High-Level Cognitive Architecture}
\end{centering}
\end{figure*}

Paul Rosenbloom and a number of other long-time members of the cognitive architectures community (which was a fairly major subset of the AI community before the recent shift toward large neural models, and is still a vibrant sub-field within academia) have synthesized numerous empirical and theoretical inputs to form what they call the ''Standard Model of Mind,'' very loosely depicted in Figure \ref{fig:sm}.  While one could nitpick at various details, on the whole I think this has been an ambitious, worthwhile and fairly successful attempt to synthesize and summarize the common elements among the understandings of human-like cognition to emerge from a diversity of cognitive psychology, cognitive neuroscience and AI initiatives.  It's reasonably interesting to look at how the Hyperon cognition approach compares to the various components identified in the Standard Model of Mind.

A brief summary of the Standard Model of Mind was given in Ben Goertzel's recent paper ''Generative AI vs. AGI'' \cite{GoertzelGenAI}, so we won't repeat that here.   We also gave there a run-down of what we see as the strengths and weaknesses of LLMs along the various key components identified by the Standard Model.   Of course, neither LLMs nor Hyperon currently excel at all these aspects, and we are conscious that it's not fair to compare the current state of LLMs versus the hypothesized future capabilities of Hyperon.   However, it is our strong (and we believe fairly educated and well-founded) intuition that it will not be possible to remedy the numerous shortcomings of LLMs relative to the Standard Model without radical revisions and additions to their architecture, whereas Hyperon already has embedded in its core architecture the ability to fulfill all aspects of the Standard Model and more.

\begin{figure*}
\begin{centering}
\includegraphics[width=15cm]{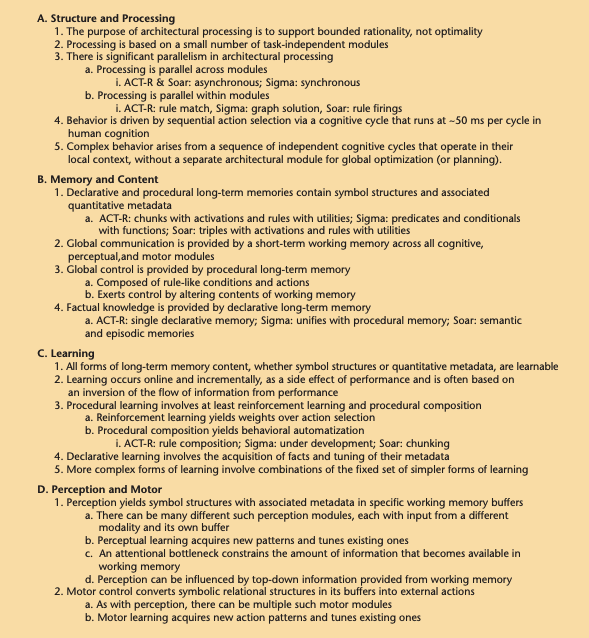}
\protect\caption{\label{fig:sma} Standard Model of Mind: Core High-Level Assumptions}
\end{centering}
\end{figure*}

We now proceed through key components of the Standard Model, explaining how they are handled in the Hyperon/CogPrime cognitive design:

\subsubsection{Episodic Memory}

Episodic memory, memory of an agent's life-history, is most centrally manifested in CogPrime via a particular ''episodic index'' that points into other memory stores such as the Atomspace and neural spaces storing sensory memories.    The nature of the episodic index is that it needs to efficiently support particular kinds of queries such as:

\begin{itemize}
\item Search for items similar to a given cue
\item Search for items matching a partially-complete cue (particularly where the completion involves elements that are physically or temporally connected to the omitted parts)?\item Search for items that were at some point physically related to a given cue
\item Search for items occurring at the same time as a given cue
\item Search for items occurring at the same or a nearby place as a given cue
\end{itemize}

One well fleshed out way to create index structures well-suited for these sorts of queries is to use hypervectors (large sparse bit vectors or integer vectors).   Hypervectors were pioneered by Pentti Kanerva in the 1980s, but like many other classical AI paradigms have begun to finally fulfill their conceptual promise in the modern era when supplied with current levels of compute power and data.  

In this approach, a Hyperon instance will contain a hypervector store that contains hypervector embeddings for Atoms in Atomspaces, as well as potentially for observed activation patterns in the neural nets within neural spaces.

The hypervector-based approach has the ancillary advantage that query processing using hypervectors can be efficiently implemented on the GSI APU chip (the subject of current experimentation), or the Simuli hypervector chip (the latter currently in prototyping phase).      

\subsubsection{Working Memory}

Working memory, in a Hyperon Atomspace as used in CogPrime, consists in effect of the Atoms with sufficiently high Short-Term Importance (STI) values.   The dynamics of ECAN can be configured so that Atoms with STI above a certain threshold are considered the ''Attentional Focus'' and subject to differently-tuned ECAN parameters.  Various cognitive processes may then be configured to select Atoms from Attentional Focus for various purposes, and Atoms from the rest of an Atomspace for other purposes.

The movement of Atoms between Attentional Focus and the remainder of an Atomspace is governed by the dynamics of ECAN as well, and may be tuned to keep the system in a reasonable attentional regime, avoiding the pathologies of obsessive-compulsiveness or scatter-brainedness.  A number of practical experiments in this regard were carried out on OpenCog Classic \cite{harrigan2014guiding} \cite{belachew2018shifting} \cite{goertzel2016controlling} .

SingularityNET Chief AI Officer Matt Ikle', who has been working on variations of ECAN in various AI systems since the late 1990s, provides the following summary of several aspects of this sort of system dynamic:

{\tt \small "One of the critical challenges confronting any system aimed at advanced general intelligence is the allocation of computational resources. Hyperon's attention-allocation system is provided by the Economic Attention Networks (ECAN) module. Founded upon an economic metaphor, ECAN serves as a key element in driving Hyperon's dynamics toward self-organizing emergence of the sorts of complex high-level network structures we believe are required to achieve Artificial General Intelligence. 

"In terms of structure, ECAN is a graph of untyped nodes, and links that may be typed as HebbianLink. Each Atom in ECAN is weighted with two numbers, STI (short-term importance) and LTI (long-term importance.) A system of equations, in which STI and LTI values are treated as artificial currencies, governs importance-value updating. These equations spread importance among various atoms within the system based upon the importance of their roles in performing actions related to the system's current goals. An important concept within ECAN is the attentional focus, consisting of those atoms currently deemed most important by the system in terms of achieving its goals. 
As we port ECAN to the Hyperon framework, we will leverage enhancements aimed at speed and efficiency. One possible enhancement would be to use the natural gradient (the gradient over probability distributions) to follow the direction of steepest descent on the space of loss functions of the parameter space. In initial experiments, such an approach displayed dramatic speed and accuracy improvements.

"We also foresee several algorithmic and architecture improvements designed to guide the process in which 'complexity emerges from simple and stable building blocks'. As described in more detail in the section on Hyperon complex dynamical systems, we envision construction of waypoints, aligned with observed natural quantities, to help guide development leading to desired emergent phenomena at multiple scales. We also plan to experiment with measures, such as Tononi's Phi, that appear related to sentience and consciousness." \footnote{ Some prototype work measuring Tononi Phi in OpenCog Classic while carrying out practical tasks is described in \cite{sellmanusing}}}

\subsubsection{Procedural Memory}

Procedural memory, in Hyperon, consists at the most basic level of MeTTa programs.   However, MeTTa is a meta-language by nature, and by implementing different type systems in MeTTa one can make it ''impersonate'' a variety of different programming languages, logic-programming and functional-programming paradigms being the most natural.   Using this facility, procedures of different sorts (e.g. cognitive heuristics for proving math theorems, procedures for controlling movements of a robot, procedures for directing the flow of a natural language dialogue) may potentially be implemented in different sub-languages represented by different MeTTa type systems.   Such types systems can initially be implemented by human programmers, but the reflective nature of MeTTa also makes it natural for type systems to originate via the system's own learning and reasoning.?
It's also key to understand the role played by appropriate abstraction in the representation of cognitively meaningful procedures.   Much of the procedural knowledge relevant to human-like cognition is vague and uncertain, and consists of higher-level processes that need to be instantiated as particular processes in a context-dependent way, using intuitive reasoning.   This means that procedure execution is sometimes straightforward MeTTa program execution, and sometimes a subtler cognitive process of mapping a vague/abstract network of Atoms into a precise MeTTa program suitable for direct execution.  

The vague/abstract networks of Atoms representing general procedures will often leverage special Atom types referring to temporal or causal relationships.  Significant prototyping work has recently been done involving temporal reasoning on Atomspaces, for instance in the context of controlling simple agents in Minecraft \cite{potapov2022minecraft}, or playing games like Pong \footnote{MeTTA for Pong is Aug 2023 prototyping work by Patrick Hammer}.

Conversion between procedural and declarative knowledge then takes the form of Curry-Howard-like correspondences (e.g. Greg Meredith's OSLF) between programs and logical expressions \cite{meredith2023oslf}.

\subsubsection{Reasoning}

Hyperon infrastructure supports a great variety of formal reasoning systems; different logics, for instance, can fairly straightforwardly be represented by different MeTTa type systems.   A variety of approaches to managing uncertainty are also supported, including fuzzy truth value algebra as well as first and higher order probabilities and various approximations thereof.

Particular attention has been paid to an uncertain reasoning framework called PLN (Probabilistic Logic Networks), which combines higher order fuzzy and probabilistic reasoning with predicate and term logic, along with human cognition motivated approaches to intensional and causal reasoning, in a formalism that explicitly grounds logical knowledge in the observations of a cognitive agent.

The biggest challenge for PLN (and every other logical reasoning system) has always been inference control -- the choice of which logical inference steps to carry out in which contexts.   A variety of heuristics can be utilized here, but we believe the core approach needs to be reflexive and history-based.   Basically: 

\begin{itemize}
\item Do some simple inferences, on some simple reasoning tasks
\item Do some pattern mining to figure out what series of inference steps have worked well in what contexts, for the simple reasoning tasks
\item Do some inference to figure out abstract principled underlying the results of the pattern mining
\item Use the results of pattern mining and inference to synthesize new series of inference steps for new problems, enabling the system to carry out slightly less simple inferences
Etc.
\end{itemize}

While this approach is conceptually straightforward, it requires significant scale in order to succeed.  This will be a key focus of Hyperon R\&D in the next phase, and it may end up a beautiful example of an AI approach that has been around a long time, and was simply waiting for more data and more processing power to really come into its own.

Nil Geisweiller, the lead developer and researcher on PLN In both OpenCog Classic and more recently Hyperon, summarizes the current state of PLN in MeTTa as follows

{\tt small There are at least two features a reasoning system geared towards AGI
must have.

1. Flexible (chaining): it needs to be able to generate inference
   trees in any direction, forward (from axioms to theorems), backward
   (from theorems to axioms), outward (from lemmas to axioms and
   theorems), inward (from axioms and theorems to lemmas), and more
   generally "omniward" (from/to axioms, theorems, lemmas,
   corollaries, etc).

2. Programmable (inference control): any computational step occurring
   during chaining should be interceptable, at a sufficient level of
   granularity, so that, given the right heuristical knowledge,
   provided or learned, the production of successful inferences can be
   made arbitrarily efficient.

Early experiments with Hyperon/MeTTa has already shown excellent
promises towards realizing the first feature.  Indeed, by combining
non-determinism and unification, which MeTTa possesses, we were able
to implement a brute force omniward chainer with just a few lines of
code \cite{nil2023temporal} .  Then use that chainer as the underlying engine of a port
in-progress of PLN.  Today, we can already do things what we were
never able to do with OpenCog Classic, such that running direct
evidence rules backward.  So with Hyperon/MeTTa we have, it seems,
already fully realized the flexibility requirement.

The requirement of a programmable inference control is expected to be
realized once minimal-MeTTa is completed, which should enable a human
or machine programmer to overwrite the reduction step (as in
beta-reduction) and interject any conditional to prune or prioritize
non-deterministic reduction.
}

Geisweiller also notes that, on the software as well as mathematical and conceptual levels, it seems to be working out in Hyperon to consider pattern-mining as essentially a special form of inference with an unusual control structure:

{\tt \small It's been already established that pattern mining can be viewed as a
specialized form of reasoning \cite{EGI2}.  The advantage of adopting such a
view is that it makes hybridizing simplistic syntactic-based pattern
mining with sophisticated semantic-based pattern mining, natural and
seamless.  An additional advantage is that inference control can be
used in both forms of reasoning, simplistic and sophisticated, to
gain more efficiency.  A proof-of-concept is currently being built
using the Hyperon/MeTTa chainer.}

Another interesting twist in the PLN/MeTTa story relates the semantics of PLN expressions to the use of comprehensions in MeTTa.   As Greg Meredith has pointed out, comprehensions can be used to concretely implement realizablity-based semantics for PLN.  The connection between this semantic approach and the presentation of PLN semantics in terms of higher order probabilistic types as alluded by Warrell's analysis \cite{warrell2022meta} appears a fertile research topic.

\paragraph{Explicit Logic vs. LLM Reasoning}

The relation between PLN reasoning and the reasoning LLMs carry out bears some note here.   It would appear that LLMs are currently capable to handle a broad variety of human commonsense reasoning, about knowledge domains well represented on the Internet.  However, they fail badly at purely formal reasoning, and also at reasoning that needs to bridge the formal and commonsensical (e.g. undergraduate physics or economics problems).  They also cannot be expected to generalize to domains with radically different properties than the data on the Internet which was used to train them.    The approach to LLM/Hyperon integration currently being pursued, as outlined above, involves leveraging the strengths of LLMs for reasoning about commonsense everyday situations along with the strengths of PLN for more general reasoning, and coupling the two together in a variety of natural ways.   We believe this will provide a viable and relatively rapid path to conquering e.g. physics and economics problems, and after that to more ambitious general intelligence.   

This approach, however, should not be construed to imply a judgment that PLN in itself could not carry out commonsense reasoning perfectly well without LLMs.   Rather, we are quite confident it can do so; however we are open to the possibility that LLMs (in current or improved form) provide a more efficient route to commonsense reasoning, because such reasoning turns out to be more about figuring out what made sense in previous similar situations than about carrying out chains of surprising or original inference steps (which is what scientific and mathematical reasoning is commonly about, and is something at which formal logic engines seem to significantly outperform LLMs).

It may also be very interesting to explore the use of LLM reasoning chains, mapped into PLN reasoning chains, as data to feed pattern mining and inference used to learn inference patterns that then guide PLN.  This will be of limited use for formal reasoning at which LLMs do poorly anyway, but may be useful to help guide PLN in mixed formal/commonsensical domains.

\paragraph{Reasoning About Space and Time}

One aspect of reasoning that has occupied significant effort in OpenCog Classic R\&D is inference regarding space and time.   As AGI researcher Hedra Seid notes,

{\tt \small "Spatial and temporal reasoning are fundamental cognitive processes that we use to make sense of our surroundings, predict outcomes, and make informed decisions in this dynamic and ever-changing world. As AI strives to achieve artificial general intelligence (AGI), it has made impressive progress in mimicking human cognitive functions. Within these abilities, spatial and temporal reasoning play key roles in understanding its surroundings and forecasting future events. Given this context, Hyperon appears to be a strong potential candidate for building such reasoning engine due to the following:

\begin{itemize}
\item Structured Knowledge Representation:
\begin{itemize} 
\item Hyperon provides a structured framework for representing concepts, facts, relationships, and rules.
\item Provides rich and expressive language that can capture spatial (such as object arrangements and orientations), temporal sequences (such as causal chains and event orders) and general knowledge relationships with precision, making it suitable for various reasoning tasks involving complex details.
\end{itemize}
\end{itemize}

\begin{itemize}
\item Type-Driven Development:
\begin{itemize} 
\item Hyperon's type-driven approach encourages structured coding.
\item Types guide knowledge representation development, fostering well-structured, readable, and maintainable code.
\item Reduce run-time errors by identifying type-related issues during compilation.
\item Allows types to rely on values, creating expressive types, and facilitates proving properties such as event overlapping and sequencing.
\end{itemize}
\end{itemize}

\begin{itemize}
\item Scalability:
\begin{itemize} 
\item Hyperon's distributed Atomspace (DAS) is designed to handle large-scale knowledge representation which is particularly advantageous for spatial and temporal reasoning, which often involves extensive datasets and complex interactions."
\end{itemize}
\end{itemize}

}

PLN's capability for spatial and temporal reasoning has been developed recently in the context of the ROCCA project, focused on using PLN to control simple agents achieving simple goals in simple environments, similar to toy domains commonly used to experiment with reinforcement learning \cite{nil2023rocca}.  As work with Hyperon progresses, these same tools will be applied in more ambitious virtual worlds like Sophiaverse, in the context of physical robot control and in a wide variety of other settings necessitating and benefiting from the greater scalability of the Hyperon framework.

\paragraph{Enabling a Hybrid Approach to Solving Technical Problems}

While LLMs have proven rather capable at many sorts of commonsense reasoning,  and at some sorts of exams like the US law school admissions test, their performance at undergraduate science exams has been much less impressive.  The Google Minerva system, specially fine-tuned for the purpose, gave a reported accuracy of around 30\% at a collection of problems drawn from MIT's OpenCourseware, problems like

{\tt \small
Each of the two Magellan telescopes has a diameter of $6.5 \mathrm{~m}$. In one configuration the effective focal length is $72 \mathrm{~m}$. Find the diameter of the image of a planet (in $\mathrm{cm}$ ) at this focus if the angular diameter of the planet at the time of the observation is $45^{\prime \prime}$.
}

\noindent or

{\tt \small
Preamble: A population of 100 ferrets is introduced to a large island in the beginning of 1990 . Ferrets have an intrinsic growth rate, $r_{\max }$ of $1.3 \mathrm{yr}^{-1}$.

Problem: Assuming unlimited resources-i.e., there are enough resources on this island to last the ferrets for hundreds of years-how many ferrets will there be on the island in the year 2000?	
}

Now these problems are clearly not easy for the average person without a fair bit of special study first.   However, they are very straightforward for a typical science student who is reading the texts and doing the homework.

Intuitively, there are four aspects to solving these problems: 1) English / common sense, 2) logical reasoning and problem solving, 3) arithmetic and algebra, 4) understanding of the domain area.

GPT4 as of September 2023 seems to be quite good at solving these problems and also some harder ones.   In fact, GPT4 with some chain of thought type prompting can even solve some considerably harder problems like \footnote{See \url{https://chat.openai.com/share/0c4070a1-5aca-4f8a-b2e0-b693efd37e53} for ChatGPT's solution obtained via chain of thought style prompting, versus failure \url{https://chat.openai.com/share/c0dd1db0-7d78-4632-8b88-758429a2dbc0} without chain of thought}:

\begin{verbatim}
An eclipsing binary consists of two stars of different radii and effective temperatures. Star 1 has radius R1 and T1 , and Star 2 has R2 = 0.5R1 and T2 = 2T1 . Find the change in bolometric magnitude of the binary, ?mbol , when the smaller star is behind the larger star. (Consider only bolometric magnitudes so you don?t have to worry about color differences.)
\end{verbatim} 

However, as reported in \cite{GoertzelGenAI}, it fails on other problems that are only modestly harder, including physics problems like

\begin{verbatim}
The motion of a star through a disk galaxy can be modeled as a point mass m released from rest at a distance d above a disk of radius R and thickness L, where L << d << R.  The disk has a uniform density and a total mass M >> m.   Describe the motion.
\end{verbatim}

\noindent and number theory problems like

\begin{verbatim}
Find all arithmetic progressions with difference of 10 formed of more than 2 primes
\end{verbatim}

\noindent These are tougher than the ones Minerva was tested on, and are more at the level that a smart advanced undergrad or grad student would be able to solve with some work, but some students with relevant background. might still fail on.

The precise boundary of what LLMs can do is a moving target and complex in nature.  E.g. it seems like just a little debugging should let GPT4 solve the above number theory problem -- but then there are a lot of other number theory problems which are more involved in various ways yet still elementary and well within the grasp of good human math students.   Even given all the subtleties, some useful generalizations can be made, one of which is that when a problem involves multiple steps of reasoning put together holistically in a "non-obvious" way, LLMs are likely to get confused.

How might we solve these problems, including the harder ones, using integrative AI in a Hyperon system?   To simplify a bit -- LLMs are good at the commonsense reasoning part, PLN inference is good at the logical reasoning part, and our "semantic parsing" initiative, aimed at parsing English into logic, provides a viable path to syncing up commonsense reasoning in LLMs with PLN reasoning.  Knowledge of relevant domain areas can be obtained by parsing a lot of English text into logical form and importing it into an Atomspace.  The only additional ingredient to these, needed to handle these problems, would seem to be some tooling for having PLN ground certain reasoning steps in calls to an external computer algebra/arithmetic system (such as Julia Symbolic, for example).  So when e.g. PLN needs to simplify an algebraic equation, or check if two equations are equivalent, or factor a number etc., it recognizes that it should invoke an external tool for this and does so.   

This is clearly not the exact strategy that humans follow, and it would be possible to teach a Hyperon system to perform in a more humanlike way -- doing arithmetic calculations via learned MeTTa procedures, or via using a robot's fingers to push the buttons of a calculator.   However, the Hyperon approach is not to try to precisely copy the way human solve each particular class of problems, but rather to use a cognitive architecture that captures the key AGI-relevant aspects of the human mind, and then have this architecture learn to solve problems in ways that leverage all the strengths available to it.

\paragraph{LLMs and Hyperon in Automated Theorem Proving}

A step beyond these sorts of undergraduate science exercises is the use of AI to guide automated mathematical theorem-proving beyond the level of elementary puzzles and exercises.   Zar Goertzel, a researcher in the use of various AI algorithms including LLMs to improve the performance of automated mathematical theorem provers, explains some of the nuances that arise in this domain making a Hyperon approach particularly appealing:

{\tt \small
"To understand the shortcomings of LLMs for automated theorem proving, even when applied in a clever and appropriately configured way, let's first look at a relatively successful application of LLMs in the. mathematics domain: The use of LLMs to generate programs that output mathematically meaningful sequences of integers, as pursued in the Alien Coding project  \cite{gauthier2023alien}. 

"For instance, the sequence $(	1, 1, 1, 1, 2, 3, 6, 11, 23, 47, 106, 235, 551, 1301, 3159, 7741, 19320,...)$ can be produced in a number of ways, e.g. the $n'th$ term is the number of trees with $n$ unlabeled nodes.   One can feed a short sequence like this to an LLM, along with a description of the meaning of the sequence -- and then ask it to come up with a program that will rapidly produce the sequence, not only the terms given but the following terms as they get larger and larger moving toward infinity.

"In this context, the first thing one generally wants to do is prove that a certain proposed program actually will generate the target mathematically meaningful sequence, for any $n$.  However,  LLMs cannot reliably do this right now.   What they will do is check that a program correctly generates the first $n$ integers of the target sequence, instead of proving that the program is correct for all elements of the sequence, which is a much harder task. Thus, such programs  as produced or evaluated by an LLM can only be considered probabilistically correct.   But this is a domain where probabilistically correct, unless the probability involved is incredibly close to certainty, is usually not good enough.   When programming or doing mathematics, messing up a single term in a moment of hallucination can create fatal execution errors or ruin a whole proof. 

"Hyperon provides a number of tools that seem  potentially valuable for improving this sort of program synthesis. The MeTTa language is designed to facilitate the evolutionary exploration of the space of programming languages or proof calculi in which to generate or search for proofs.  And it is also designed to work natively with probabilities and other ways of gauging uncertainty.   If working with proof terms that only probabilistically represent what we wish them to, then we will wish to use probabilistic logic to reason about them, a function for which Hyperon's PLN framework is ideally suited.

''In general, the largest obstacle blocking LLMs from successfully coming up with proofs in this or other domains is their difficulty in chaining together multiple steps in a context-appropriate way.  Memory is also an issue as one may wish to hold some parts of a proof in mind while working on others, which LLMs currently have no clear mechanism for, but a broader Ai framework such as Hyperon can facilitate.   Interestingly, these two shortcomings also hold back traditional automated theorem provers in different ways; they also struggle with strategic construction of lengthy inference chains (which is why interactive theorem proving is so common) and with using various sorts of memory to guide choice of inference steps.   Hyperon's more cognitive approach has some promise for overcoming these limitations.

"Hyperon also has potential for bridging the informal proof domain where LLMs have most promise, and the world of extremely detailed, low-level proofs where typical automated theorem-provers live.   In Hyperon, the high-level proof sketch (similar to what mathematicians might outline) can be translated into formal logic for automated symbolic reasoning. This way, the strengths of LLMs and automated theorem provers can both be synergistically leveraged."
}

\subsubsection{Reinforcement Learning}

Reinforcement learning, conceptually, is a matter of rewarding (increasing attention to and odds of selection in similar contexts of) procedures that have led to good results, and punishing (decreasing...) to procedures that have not led to good results ... and choosing new procedures to try by varying appropriately on those that have been successful in the past.

In this sense, a combination of ECAN with PLN (applied to logical versions of executable procedures) would be described as implicitly doing reinforcement learning, for example.   Augmenting these methods with probabilistic program synthesis, and explicit induction of probability distributions from the set of procedures tried already, would bring something a bit closer to standard reinforcement learning algorithms.

It would also be possible to implement classical RL algorithms in MeTTa and run them on Atomspace; however, these have well-known issues dealing with complex real-world situations in which there is no reward function tied in a simple way to agent actions, or in which a key role is placed by subtle and multidimensional intrinsic rewards.  

Robot actuation would seem to be a case where classical RL has a natural role to play.  The interfacing of high level action planning with low level physical movement planning might then take the form of probabilistic programming involving induction of probability distributions spanning physical movements and action plans, where the probabilistic program synthesis would then guide both the search in the classical RL algorithm guiding movements and the search in the more abstract ''reinforcement learning like ECAN/PLN approach'' guiding planning.

\subsubsection{Language Learning and Usage}

LLMs have shown tremendous proficiency in natural language dialogue and also at a number of the standard tasks from the computational linguistics field.   However, they demonstrate severe deficits in semantic understanding, especially in contexts where formal knowledge plays a role; and their linguistic productions lack subtlety and aesthetics and the ability to compel.   

As NLP researcher Andres Suarez notes, {\tt \small  "While the advent of Large Language Models (LLMs) has marked significant progress for natural language processing, the field still faces significant challenges. The Hyperon architecture, designed to integrate seamlessly with LLMs, offers a promising avenue for overcoming these obstacles.

"Comprehension and generation challenges in NLP include

\begin{itemize}
\item Ambiguity: A single sentence can have multiple interpretations based on its context; current models still can have trouble in complex situations that may be straightforward for humans.
\item Emotion and Tone: Accurately and consistently detecting underlying emotions, sarcasm, or humor in language is difficult for current models.
\item Dynamic Language: Language is ever-evolving, and models that cannot adapt risk becoming obsolete.
\item Contradictory statements: Observed especially when creating long pieces of text or balancing multiple pieces of information.
\item Faulty reasoning: While it's currently possible to generate text that appears logical on the surface, it can contain logical leaps that don't hold up under scrutiny.
\item Failure to incorporate information: Caused by a finite context window, and the inability to access an external database for real-time information.
\item Hallucinations: Lacking factual correctness in generated content.
\end{itemize}

"Hyperon-based solutions to these issues could look like:

\begin{itemize}
\item Integration with LLMs: The linguistic fluency inherent in LLMs can be synergistically combined with Hyperon's structured reasoning capabilities for a more holistic approach to NLP.
\item Rich Knowledge Graphs: Hyperon's architecture allows for the storage of extensive contextual information, thereby reducing ambiguity and enhancing overall coherence.
\item Grounded Reasoning: The ability to connect language models to real-world data, context, or sensory input can significantly improve both comprehension and generation.
\item Logic Engine: Hyperon's Probabilistic Logic Network offers a framework for the explicit handling of ambiguous or incomplete information, allowing for more robust conclusions.
\item Continuous Learning: Hyperon is designed for adaptability, enabling it to evolve with the language and capture emerging nuances.
\item Feedback Mechanisms: The architecture includes feedback loops that facilitate continuous learning and refinement in NLP capabilities.
\item Fact-checking: Hyperon can cross-reference facts with its extensive knowledge base to ensure the accuracy of generated content.
\end{itemize}

"Overall, the fusion of deep learning techniques with symbolic structured reasoning, as exemplified by the Hyperon architecture, holds the promise of revolutionizing NLP. This integrated approach not only aligns well with the capabilities of modern LLMs but also sets the stage for the development of NLP systems that understand and generate human language with unprecedented nuance, coherence, and contextual relevance."}

One way to look at many of Hyperon's potential contributions here is in terms of the combination of statistical approaches to NLP (of which LLMs are the preeminent examples today) and more formal-linguistics approaches.   As Ben Goertzel notes, {\tt \small "Formalization of syntax, semantics and pragmatics has a long history; however, it's also become clear that compact lists of formal structures lack the richness and subtlety to encompass natural language.   What can be done by putting LLMs and Atomspace together is to create large-scale bodies of formal linguistic knowledge, including highly particular linguistic rules and also a hierarchy of abstractions and generalizations.    This should allow close interfacing between declarative knowledge formed by PLN and other Hyperon cognitive mechanisms such as concept blending, and linguistic patterns and structures.   This could be done in a variety of ways; one promising avenue is to leverage variations of the Word Grammar formalism, which leverages graph-based and logic-based structures for both syntax and semantics.

"One can then envision a comprehension pipeline wherein LLMs are used to translate natural language sentences into a combination of logical content and formal (e.g. Word Grammar) syntactic content; and a generation pipeline wherein logical content is translated into syntactic content using formal syntactic content as a partial intermediary, thus providing an understanding of language going beyond surface-level syntactic corpus analysis and given potential for more profound linguistic inventiveness."}

\subsubsection{Multimodal Perception}

Multimodal perception, in a sense, is currently handled fairly well via deep neural networks; however, the sense in which these networks really understand what they are perceiving is fairly limited, which presents a challenge in terms of using them as the perceptual cortices of an integrated cognition system.

The natural solution to this conundrum is to build explicit links between concepts and relationships in Atomspace with concepts and relationships as represented in the weight and activation patterns in neural networks.   This could be done in various ways; the most straightforward approach would be simply to learn linear or nonlinear functions mapping specific Atoms into combinations of neurons in the networks inside neural spaces.     So for instance, if a Hyperon system sees a number of images labeled with the word ''cat'', it can then learn a mapping from the Atom representing the word ''cat'' to a combination of neural in the visual neural space.   

The next step is inductive learning of the probability distribution characterizing a large ensemble of such learned mappings.  By probabilistic synthesis based on this distribution, new mappings can be inferred in cases where there is scant or even no labeled data.

\subsubsection{Action Learning and Coordinated Action}

Action learning in CogPrime is not fundamentally distinct from other sorts of learning.   It can be carried out via RL-like methods as mentioned above; or it can be carried out more purely by PLN inference or probabilistic program synthesis not tied to any particular reward function, ''merely'' the pursuit of concise generalization of existing procedural knowledge.

\subsubsection{Goal Refinement and Goal System Management}

Hyperon is designed to be usable for strongly goal-driven systems with a great variety of top-level goals, self-organizing networks with no intrinsic notion of goal, or intermediate cases like systems that are substantially but not entirely goal-driven.

Explicit pursuit of goals is naturally done by PLN, by RL-type methods, or by probabilistic procedure learning or evolutionary procedure learning.   These methods can quite naturally balance multiple different goals on different time-scales, and even goals that contradict each other to a significant extent.  Explicit reasoning about inter-contradictory goals is also relatively straightforward using paraconsistent logic systems, which have a natural mapping into PLN's uncertainty semantics.

Top-level goals for a Hyperon system can be provided by human programmers; another option however is that they can be created via concept formation heuristics and ECAN.   A combination of these two approaches is likely appropriate.   Creation of subgoals from existing top-level goals is then a relatively straightforward application of concept formation and PLN inference.

\subsubsection{Reflexive Self-Understanding}

The Atomspace is explicitly designed for reflexive self-understanding, in the sense that MeTTa, PLN, pattern mining and other standard Hyperon/CogPrime cognitive operations are explicitly designed to act on the Atomspace metagraph as input data, as a place for output, and as a substrate for storing interim results.   This underlying design does not, of course, automatically enable reflexive self-understanding at a deep level; it just means that, 

\begin{itemize}
\item If the learning and reasoning algorithms are smart enough, there are no obstacles to very high levels of reflective self-understanding
\item The amount of intelligence needed on the part of the learning and reasoning algorithms to achieve modest levels of reflexive self-understanding, is not necessarily that high (because the Atomspace design doesn't place obstacles in the way of conceptually simple problems of self-understanding being actually simple in practice)
\end{itemize}

This suggests it may be plausible to implement a virtuous cycle of the form: A bit of reflexive self-understanding makes the system a bit smarter, which enables it to achieve a bit more reflexive self-understanding, etc.

\subsubsection{Modeling and Understanding of Other Minds}

Human understanding of other minds appears to be a combination of multiple factors, including (uncertain) logical modeling of other minds, empathic synchrony, and internal simulation of other minds.   Hyperon should be able to do all these tricks (see comments on emotion below), and in the case of simulation should be able to significantly out-do humans in some ways.   

A Hyperon instance can create other Hyperon instances specifically intended to model specific other minds, and train and tune them accordingly.   Already using LLMs we can make an interesting, though not fully accurate, ''textual twin'' of a specific human being by fine-tuning an LLM on textual data they have produced.   Emulating the face and voice of a human with eerie accuracy is also possible with current deep neural net technology.   What happens when we tightly connect these networks with an Atomspace representing the knowledge and personality of a particular person?   The TWIN Protocol project is moving in this direction with the goal of making commercial products with practical utility; however the same approach could potentially be very valuable for enabling a Hyperon instance to make inferences about what specific other minds are doing.

And the next step, of course, is to carry out inferential generalization based on these individual-mind-modeling Atomspaces.   In this way, it seems Hyperon systems may be able to more than compensate for their likely shortcomings (relative to human standards) in emotional synchrony with humans.  And when it comes to Hyperon instances modeling each other, it seems clear their ability to construct approximate simulations of each other will serve them in extremely good stead.

\subsection{What Is It Like to Be a Hyperon?}

We next consider a few core aspects of human mind that are not explicitly called out as key aspects of the Standard Model of Mind, but are considered in the Standard Model in various ways, and commonsensically appear key to human-like general intelligence.   These aspects lean a little further toward the experiential and subjective aspects of general intelligence -- what is it like to be a Hyperon?

\subsubsection{World Modeling}

The ''world model'' of an AGI system does not have to be a distinct component as it is e.g. in some robot control systems, it can be implicit across a variety of the system's knowledge stores.

However, implicit world models can still vary widely and can be more or less sophisticated and more or less useful.  It is clear for instance that LLMs have a badly insufficient world model except in some particular cases; the root cause here seems to be that their knowledge is overly particularized, consisting mainly of a vast number of special cases.   They do have some implicit abstractions in their knowledge, but their ability to adaptively deploy these abstractions is relatively minimal compared to their ability to find and morph appropriate particularities.  

A Hyperon system's world model, to some extent, consists of the abstractions it has learned from particular cases via methods like PLN and concept creation.   The learning of explicit causal  relationships is an important part of this.   RL type methods also can play an important role in learning causal relationships, often of a more concrete character, but also subject to inferential abstraction.

Computational systems also possess capabilities for world modeling not open to biological brains.  For instance, a Hyperon instance can model.a physical system by explicitly running a physics engine; it can model a computer network that's part of its infrastructure by explicitly running a network simulation, etc.   It can tune the parameters of these simulations to match its abstract understanding, and/or run these simulations under various conditions to learn new potentially relevant abstractions, etc.  This illustrates the approach wherein, while Hyperon's core cognitive architecture is to a significant degree inspired by (human) cognitive science, fundamentally (until we get to OpenCog Tachyon anyway) it is a digital-computational system and our inclination is to allow it to leverage whatever advantages its digital underpinnings allow.

\subsubsection{Emotion}

\begin{figure*}
\begin{centering}
\includegraphics[width=15cm]{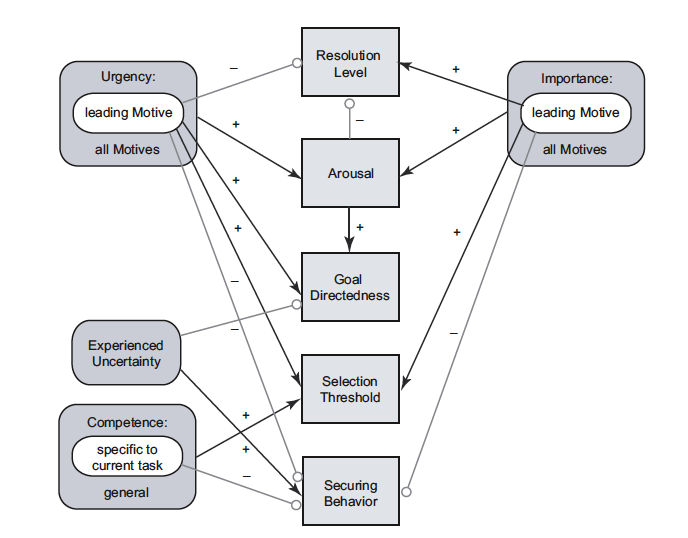}
\protect\caption{\label{fig:PsiEmotions} Graphical illustration of the basic dynamics of human-like emotion according to Joscha Bach's Psi model}
\end{centering}
\end{figure*}

\begin{figure*}
\begin{centering}
\includegraphics[width=15cm]{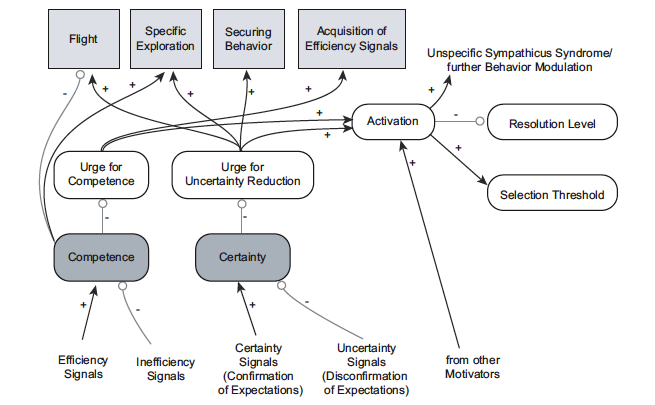}
\protect\caption{\label{fig:PsiModulators} Graphical illustration of the process via which various parameters ("modulators") guide human-like action and emotion according to Joscha Bach's Psi model}
\end{centering}
\end{figure*}

Joscha Bach's MicroPsi model of human-like intelligence  \cite{Bach2009}(depicted roughly in Figures \ref{fig:PsiEmotions} and \ref{fig:PsiModulators}), drawing on Dietrich Dorner's earlier Psi model \cite{Dorner2002} grounds the common human emotions in terms of parameters of cognitive subsystems concerned with action, perception and memory.   Cai Zhenhua's 2011 PhD thesis  \cite{Zhenhua2011}explored the implementation of this model in OpenCog Classic, in the context of simple virtual agents exploring a 3D simulation world.   We have also connected this model with Scherer's Component-Process Model of emotion \cite{scherer2009dynamic}, in the context of guiding implementation of emotion models for humanoid robots and avatars \cite{GoertzelDuncanEmotion2023}

While in some ways crude and oversimplified, we believe this line of development

\begin{itemize}
\item Adequately explains the connection between emotions and other aspects of a cognitive system
\item Gives a way to supply AGI systems with rough analogues of standard human emotions (which will only sometimes be something one wants to do, depending on what kind of system one is building)
\item Gives a way to think systematically about AGI emotions that may differ in substantial ways from the standard human emotions
\end{itemize}

\subsubsection{Creativity}

Creativity in a complex cognitive system arises from a great variety of sources.   Goertzel's book From {\it Complexity to Creativity} \cite{Goertzel1997} analyzes the roots of creativity in terms of underlying dynamics such as

\begin{itemize}
\item Varying on individual forms that have been perceived before (''mutation'' in evolutionary learning parlance)
\item Combining aspects of forms that have been seen before (''crossover'')
\item Complex, chaotic self-organization among interacting elements (''self-organizing emergence'')
\end{itemize}

The effective implementation of each of these can involve a fair amount of subtlety, however.   LLMs and other current generative AI models achieve their limited degree of creativity by varying and combining surface-level patterns observed in their training data.   Their limitations are not, it seems, mainly due to limitations of the operations of mutation and combination but rather due to the lack of easily manipulable abstractions in these systems.   Varying and combining in a more fundamentally creative way would require more abstract representations -- what Douglas Hofstader would have called more inventive and abstraction-savvy ''knob creation'' rather than ''knob-twiddling'' of adjustable features defined at the surface level \cite{Hofstadter1995}.

Creativity in CogPrime, in an art, literature, mathematics, science, or social or self analysis context, will be a combination of multiple methods including evolutionary learning, uncertain inference, probabilistic synthesis and autopoietic ''Cogistry'' rewrite-rule networks.   Creativity will be intrinsically emotion-driven, according to the underpinning of emotion in parameters of perception, action and memory articulated in the Psi model.   But above all creativity will, when effective, be heavily guided by appropriately abstracted representation of the diverse knowledge content used as raw material for new creations.

\subsubsection{Consciousness}

''Consciousness'' is a massively overloaded term, and can mean numerous things including e.g.

\begin{itemize}
\item Explicit world and self awareness of the sort that we seem to have while awake but not while deeply asleep or under anesthesia
\item The sort of reflective self-understanding that people have much more of than dogs, dogs have more of than worms, adults have much more of than babies, etc.
\item Qualia -- the raw, subjective feeling of being aware
Basic awareness/responsiveness to the environment of the sort that arguably every elementary particle displays
\end{itemize}

There are debates about whether digital computer systems can ever be conscious in the sense that people are, which are difficult to resolve scientifically given that the connection between conscious experience and physical/biological activity is not well scientifically understood yet even in humans.   Ben Goertzel has proposed using brain-brain and brain-computer interfacing to investigate these issues in new ways\cite{GoertzelObsoleting2015}.

Goertzel's paper \cite{Goertzel2014cons} explores issues regarding consciousness and the structures and dynamics in human brains and AGI systems.   From a practical perspective, however, there are so many unknowns here that the nitty-gritty work of building AGI systems tends to set aside questions of conscious experience, and instead work in terms of ''neural and cognitive correlates of consciousness.''

In these terms, the key correlate of ''reflexive self-aware consciousness''  -- the kind of consciousness that is there when awake but not deep asleep -- in Hyperon systems systems is the Attentional Focus.   The contents and dynamics of the Attentional Focus are likely the primary determinants of a Hyperon system's ''state of consciousness'' in the commonsense interpretation.

\subsection{Hyperon as an Infrastructure for Alternate Cognitive Architectures}

While the main thrust of Hyperon-based AGI R\&D, as being pursued by the team developing the Hyperon infrastructure itself, is centered in the vicinity of the CogPrime cognitive architecture reviewed above, there is also an intention for Hyperon to be useful for experimentation with alternative AGI paradigms and architectures.   

Of course, one can't expect the same infrastructure to be useful for every possible AGI approach; however, to the extent that multiple AGI approaches can be pursued within the same infrastructure and toolset, it may be easier to compare their results and strengths and weaknesses, and to combine modules, algorithms or representations that have broad utility beyond one particular AGI approach.  Along these lines, for instance,

\begin{itemize}
\item There have been discussions with members of the NARS community about the potential benefits of MeTTa and Hyperon as an implementation fabric for a version of the NARS reasoning system and cognitive architecture
\item There have been discussions about use of Hyperon to implement highly biologically realistic neural networks, perhaps using nonlinear-dynamical neurons (based on Hodgkin-Huxley equation or Izhikevich's chaotic neuron) and/or perhaps using Alex Ororbia's predictive coding based alternative to backpropagation \cite{ororbia2022neural} and/or Yi Zeng's BrainCog architecture \cite{zeng2022braincog}
\end{itemize}

\subsubsection{Hyperon as an Infrastructure for SISTER}

As another potential example of the use of Hyperon to implement and explore alternate cognitive architectures, Deborah Duong, CTO of Rejuve Network:

{\tt \small  "The SISTER (Symbolic Interactionist Simulation of Trade and Emergent Roles) framework \cite{duong2004sister} that I've been working on for quite some time now provides a promising approach for integrating LLMs with symbolic reasoning systems in a neurosymbolic architecture.

"The key advantage of SISTER is its ability to model the social emergence of symbols and meaning from lower-level dynamics, akin to how concepts arise through human interaction and collective sensemaking. As such, SISTER can generate the implicit representations that come before explicit symbolic formulations. The autonomous agents in SISTER self-organize to create shared signaling systems and conceptual spaces, which capture new abstractions as they emerge.

"SISTER provides selective pressure for communications to be both compact, as they are resource constrained, and free of context enough so that receivers with different contexts can still understand them, and composable so that things never seen before can be expressed..  It so happens that compactness and being free of contexts and composability leads to the creation of mathematics and symbol processing, Thus using SISTER for neurosymbolic knowledge extraction has an advantage over methods that pull symbols from internal neural states without selection for compactness, context freeness and composition.  

"Once these implicit representations are externalized, probabilistic logic networks like Hyperon can then translate them into explicit logical constructs understandable to humans. This allows for an expansion of knowledge through induction of new concepts, deduction through logical reasoning, and abduction of new hypotheses. In essence, SISTER provides a pathway for bootstrapping the neurosymbolic capabilities of Hyperon through emergent symbol grounding, complementing Hyperon's strengths."}

\subsection{Hyperon as a Foundation for Superintelligence}

Emulating human-like cognitive architecture at a certain level of abstraction is, we believe, a viable path toward creating human-level AGI -- even if the underlying components of the architecture don't involve simulation of biological structures and dynamics at the realistic or notional level.   However, we have already seen above that, in many cases, the way Hyperon will address concrete problems within its roughly human-like cognitive architecture is going to be very different from how humans address problems.   Humans can't wire their brains into Julia Symbolics and they can't systematically pattern-mine on their whole knowledge-bases and they can't map their learned procedures into declarative form using simple formal transformations -- Hyperon can do these things and there's no reason it shouldn't, so long as it can do so in ways that build toward rather than detract from its fundamental self- and world- understanding.

These same non-human aspects of Hyperon are likely to be key to the transition of Hyperon from human-level to radically superhuman general intelligence -- which we believe is a near-inevitability if we do succeed at getting it to human-level general intelligence.   Hyperon will be able to introspect and analyze over its whole knowledge base, and use this understanding to explicitly revise all its knowledge and rewrite or redesign all its cognitive procedures.  This is a level of self-understanding and cognitive sophistication totally inaccessible within the human brain architecture, and palpably leads on to various forms of superhuman intelligence that are frankly hard for us to grapple with even conceptually.

\begin{figure*}
\begin{centering}
\includegraphics[width=15cm]{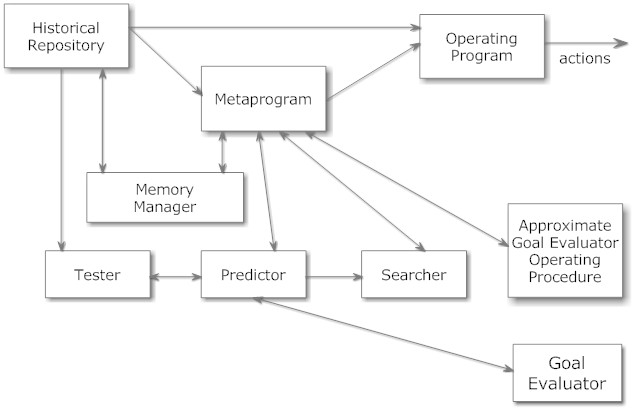}
\protect\caption{\label{fig:golem} Sketch of GOLEM meta-architecture for relatively safe/reliable self-modifying AI \cite{Goertzel14a}}
\end{centering}
\end{figure*}

The GOLEM meta-architecture depicted in Figure \ref{fig:golem} provides one very simplified way to think about the structure and dynamics of AGIs that are able to rethink and revise their own sourcecode.  In this approach there is a "base operating program" AGI system that carries out various actions in the world, and then a meta-level optimization system that searches the space of possible architectures for this base AGI system in accordance with high-level goals, and modifies the code of this base-level system accordingly -- doing systematic tests before updating the code of the base system, and also with the possibility of rolling back if things don't work.   This sort of approach could be taken in a rigid way that attempts to maintain certain initially fixed goals through all the self-modifications; or, more interestingly, it could be utilized in a more open-ended manner in which the top-level goals are allowed to modify via experience, but are usually changed only at a significantly slower pace relative to the changes in the base AGI system.   The dynamics of goal evolution as related to overall cognitive evolution in this sort of system is not at all well understood at the moment.   

As Ben Goertzel notes,

 {\tt \small "Some theorists have argued that strongly self-modifying systems like 'GOLEM with a Hyperon initial base and ability to update its goals'  would be likely to converge into pathological goal systems such as megalomania, wireheading or total self-centeredness -- but there seems no rational reason to believe this and my own intuition is quite otherwise, I suspect if there is a tendency of this sort of system to converge to some sort of attractor it's more likely one with a compassionate and beneficial nature.  But we are definitely here in a domain where current science barely points us in some directions to look, and our thinking is necessarily guided more by intuition than by rigorous results."}

\begin{figure}[htb]
\centering
\includegraphics[width=6cm]{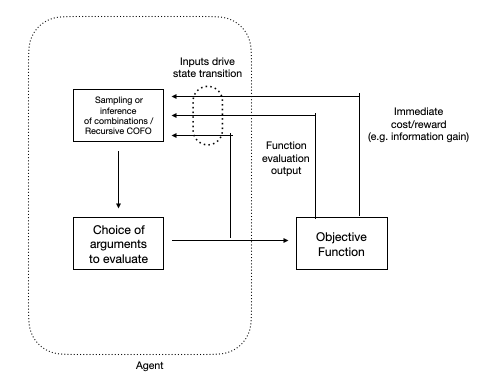}
\protect\caption{\label{fig:cofo} Schematic diagram of the Combinatory-Operation-Based Function Optimization (COFO) process for optimizing functions via searching spaces of combinations.   COFO processes follow the general template of a Discrete Decision System that chooses actions aimed at achieving goals; but the actions they are concerned with are actions of providing an argument to be evaluated by a partially-unknown function that is the subject of optimization activity.   The core algorithms involved in AGI systems like OpenCog can be expressed as cases of the COFO process.}
\end{figure}

The ''General Theory of General Intelligence" articulated in \cite{GTGI} gives some reasonable tools for thinking about the spectrum of systems between something like a ''Hyperon 1.0'' and a mature GOLEM system.   GTGI's formulation of cognitive algorithms in terms of Galois connections leverages a representation of these cognitive algorithms as special cases of a "Combinatory-Operation-Based Function Optimization" (COFO) process as roughly depicted in Figure \ref{fig:cofo}.   This sort of process can be carried out with great precision if one has closer-to-infinite compute resources, and then in various more or less heuristic ways if one has currently-realistic compute resources -- where human-like intelligence then corresponds to a particular network of heuristics that are especially valuable in historically human-relevant environments.   Looking at how COFO works when one has tremendously but not insanely more compute resources is interesting and could give insights into potential continuous paths of evolution between HCAGI and radical ASI.  This relates to the thoughts in the article {\it Robust Cognitive Strategies for Resource-Rich Minds} \cite{goertzel2022robust}, which explores cognitive strategies like

\begin{itemize}
\item Statistically sampling from neighboring possible worlds in the multiverse.
\item Maintaining a history of your prior versions, assessing their view of your current self and rolling back now and then when appropriate
\end{itemize}

\noindent that may well be possible for physically-feasible post-Singularity superminds but are beyond the scope of basic HCAGI cognitive heuristics.

There are of course many steps to get to this sort of place in reality -- we are all well aware that current pre-alpha Hyperon systems lack many aspects of general intelligence observed in ordinary one year old children.   However, it is also important to have as clear as possible of an understanding of the broader potentials of the architecture we are working with.

\subsubsection{Hyperon and Universal AI Methods}

While it's likely superintelligence will break all human theoretical and intuitive concepts in many ways, it also seems likely that our mathematical theories may encompass aspects of superintelligence that go beyond our everyday intuitions.   The mathematical theory of AGI contains approaches like Marcus Hutter's AIXI and Juergen Schmidhuber's Godel Machine that are unrealizable using currently realistic compute resources, but may capture some aspects of how radically superhuman minds may think.  Alexey Potapov has commented on the relationship between Hyperon and these mathematical approaches to general intelligence as follows:

{\tt \small
"Hyperon as a platform doesn't contain implementations of universal induction or AIXI and doesn't even account for information-theoretic criteria in its core design principles. Nevertheless, its development was motivated by the information-theoretic approach to AGI as well.

"Pattern matching is a basic operation in Hyperon, and all algorithms are represented in the form of such patterns as well. This makes Hyperon friendly for working with algorithmic regularities, which are represented in a declarative composable form. One particular Hyperon module is Pattern Miner, which uses information-theoretic surprisingness criteria. Thus, while Hyperon doesn't insist on implementing AGI systems on top of universal induction, it facilitates the use of elements of algorithmic information theory.

"Turing-complete probabilistic programming languages (PPLs) can be considered as a practical implementation of universal induction and related theories. Indeed, it is enough to write a probabilistic program, which generates all other programs, and condition it on the observations to get universal induction approximated by a specific sampling or inference algorithm implemented in the PPL interpreter. Efficient inference in PPLs corresponds to efficient universal induction. But it might not be achievable in AGI settings without knowledge-based declarative reasoning, meta-heuristic search (e.g., genetic programming), etc. No programming language or framework provides tools for implementing such inference control over universal probabilistic models, while it is one of the running use cases for Hyperon.

"Some models in the universal artificial intelligence domain are heavily based on logic. Godel Machine is one such example. Some recent variations of AIXI also suppose extraction of ''laws of nature'' in a declarative form amenable for reasoning. Implementing Godel Machine with its self-rewritable axiom systems in Hyperon might be easier than in any other framework.
 
"Thus, while Hyperon is not built around universal induction and AIXI, it can be useful for experimenting with their different extensions going far beyond basic models, and possibly for developing a full-fledged solution on their basis."
}

As Ben Goertzel notes, {\tt \small "In 2008 when I first met Nil Geisweiller, who has become the lead AI developer behind PLN and metagraph pattern mining and a number of other aspects of Hyperon AI, and described my ideas about AGI design to him, he said he thought it was essentially an approximative Godel Machine (referring to Juergen Schmidhuber's theoretical idealized AGI system that, loosely speaking, carries out actions and self-modifications based on what it can formally prove is most likely to work out.).  I did agree with him, but I also felt of course that this was sort of faint praise of my elaborate cognitive architecture thinking.  There are much simpler approximations to the Godel Machine, but finding an approximation that will work well for the tasks humans care about, running on reasonable available resources, is basically just another way of phrasing the whole problem of cognitive architecture design.   

"A few years later I articulated what I now call the 'Mind-world Correspondence Principle' -- i.e. that the way to make a system that has a reasonable degree of general intelligence relative to a certain set of goals in a certain set of environments, is to embed in the structure of this system a set of patterns that are closely homomorphic to patterns in the environments relative to the goals.  So for instance, if an environment is hierarchical and the system's goals tend to break down at least approximately into subgoals pertaining to different levels of the hierarchy, then to efficiently deal with this situation, the mental architecture of a cognitive system should probably have distinct hierarchical patterning as well. 

"A generally intelligent meta-system like GOLEM would be able to identify the patterns in its current environment /goals and goals and then modify its operating program to better reflect these patterns.   Less thoroughgoingly self-modifying systems like human minds and cultures still do something similar to some extent, as they adapt and develop."}

Geisweiler still takes a Godel Machine related perspective on his Hyperon AGI work:

{\tt \small

A Goedel Machine \cite{Schmidhuber2006} is an ideal, but realizable, self-rewriting
system which only triggers rewrites that have been formally proven
to improve the system.  The difficulty comes from generating such
proofs.  Any long chain of reasoning, which such target theorems
likely require, is extremely hard to come by.

Inference control meta-learning may provide a way to remedy that.  By
recording and mining traces of past successful and unsuccessful
inferences, inference control rules can be discovered and used to
speed-up future reasoning by biasing the search towards inferences
that have been deemed successful in the past.  The problem however is
that in order to learn to discriminate success versus failure, one
needs to experience success, at least once.  In the context of finding
proofs of self-rewriting improvements, this may be difficult.

Things can be done about that nonetheless.  First, some rewrite may be
sufficiently simple, like tuning some parameter, such that some proof
could be found within some acceptable amount of time.  In that manner,
a corpus of increasingly difficult proofs could conceivably be built.
Second, and that is a deeper point, some knowledge about inference
control for self-writing improvements could be sufficiently general
that it may be learned from solving problems that do not pertain to
self-rewriting improvements.  The extend to which this is true remains
to be seen, but if that is the case, even a bit, then it would provide
an effective way to kickstart a virtuous cycle of self-improvements,
as by merely learning to solve problems in the world, the system would
learn to solve problems within itself as well.

}

\subsubsection{Complexity, Self-Organization and Emergence on the Path to Superintelligence}

The initial version of Hyperon is being designed with great care, and we have explored many of the underlying design principles and ideas here.   However, if the project succeeds, as Hyperon versions surpass the human level in general intelligence, they will fairly rapidly become self-designing, self-modifying systems -- placing them even more squarely in the domain of complexity science than they will be at the earlier pre-human-level stage, when "only" their knowledge and cognitive heuristics are complexly evolving but their core algorithms and meta-representations will be fixed.

This suggests that, alongside notions from Universal AI theory, notions from the science of chaos and complexity may end up being valuable for understanding later stages of Hyperon (self-)development.   As Matt Ikle' notes,

{\tt \small "Artificial General Intelligence (AGI) research is riddled with many large foundational and definitional questions. What are intelligence, consciousness, and life, after all? How did the first single cell organism form? How can one create ''something from nothing''? How can one take silicon, alter its electrical properties, add electricity and a fair bit of programming, and end up with AGI?

"Partial answers to these questions may lie in the large class of ''chaos-theory'' related disciplines such as nonlinear phenomena, self-similarity, fractals, complex dynamical systems, self-organizing maps, self-modifying systems, phase transitions, and emergent phenomena. 

"The importance of such nonlinear phenomena for creating AGI appears to be born out in current neuroscience research. In a recent week-long neuroscience experiment (XXhttps://www.ncbi.nlm.nih.gov/pmc/articles/PMC10081438/), human EEG dynamics were described as 

{\it "patterns of ''punctuated equilibrium'': periods where networks would remain in stable states that corresponded to behavior and were interrupted by transitory bursts that were difficult to predict, displayed chaotic characteristics, and coincided with behavioral transitions. Brain state statistics displayed power laws characteristic of critical dynamics that are a trait of systems where complexity emerges from simple and stable building blocks. These results indicate that the complex and flexible brain dynamics that underpin real-world behavior are an emergent property of mixtures of individual, stable networks with simple dynamics.}

"Such punctuated equilibrium activity patterns are remarkably similar to the strange attractors of the complex dynamical systems theory underpinning Hyperon.  At first these attractors will pertain to knowledge gained by the system within the framework of human-supplied cognitive algorithms; then later they will pertain to the formation and modification of the system's cognitive algorithms and low-level implementation code as well."
}

Complex systems theory was also one of the main inspirations for the theory of Open-Ended Intelligence mentioned above, which views general intelligence as a self-organizing process combining dynamics of individuation and self-transcendence.    In the OEI perspective, reaching toward Universal Intelligence is something that general intelligences do in an effort to exceed their limitations and self-transcend, and an effort to outsmart conditions that threaten or weaken their individuation.   Understanding what happens as OEIs reach further and further toward universal intelligence in various ways is important, but is only one among many interesting aspects of the complex self-organizing emergent dynamics of the evolving and overlapping intelligences that comprise the most fascinating parts of our world.  In this view, the success of a system like Hyperon as an Open-Ended Intelligence will come if the system survives for a reasonable period of time, and while doing so, transforms itself into a radically broader and richer kind of system than anything its human creators or its initial version could possibly have conceived.

\section{Hyperon's Development Path}

Having roughly described what we are aiming to build, how do we plan to do it organizationally?

\begin{figure*}
\begin{centering}
\includegraphics[width=15cm]{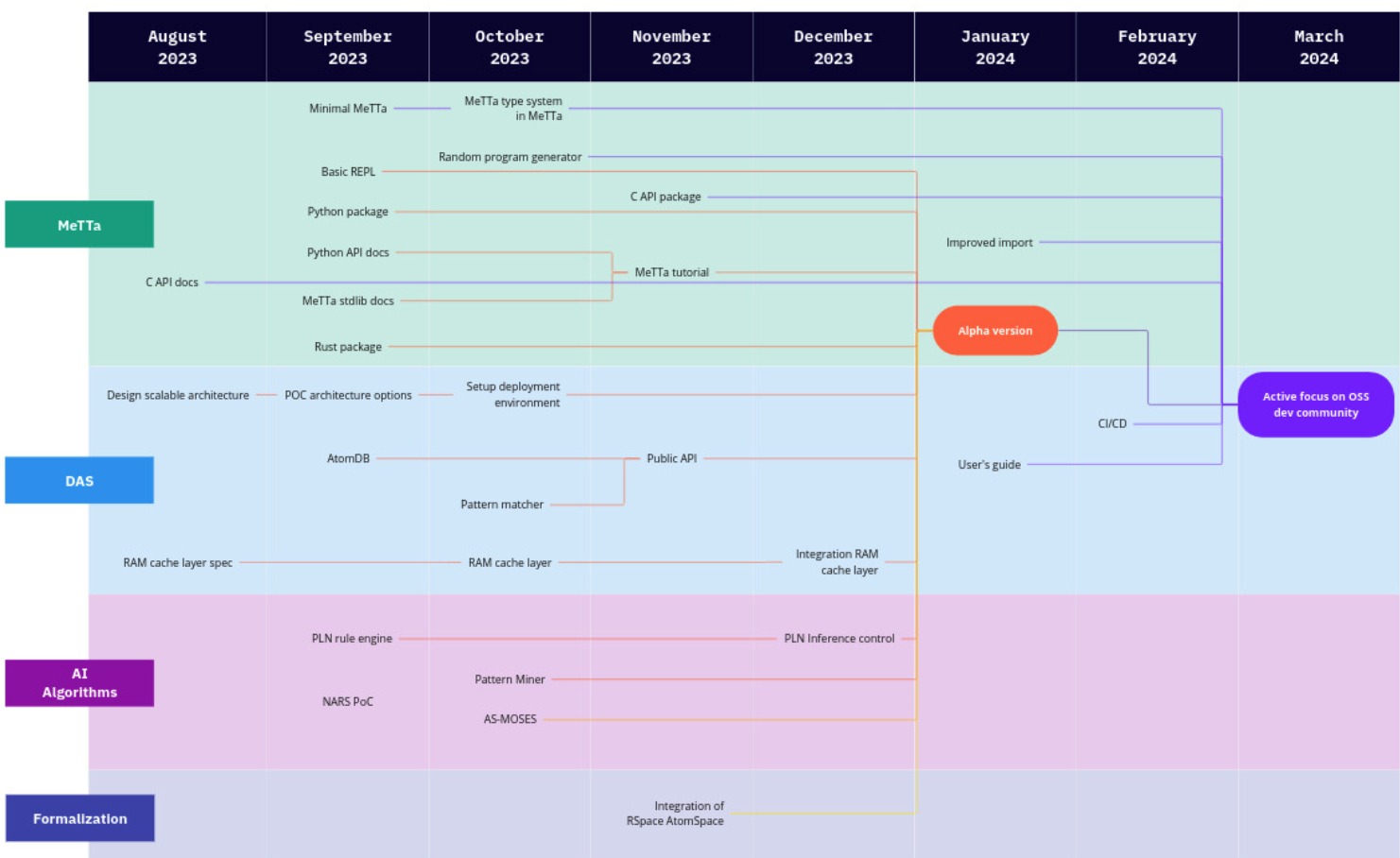}
\protect\caption{\label{fig:Hyperon-Roadmap} Rough tentative roadmap for some of the key development initiatives regarding core OpenCog Hyperon technology (not attempting to cover specific applications, demos or commercialization efforts)}
\end{centering}
\end{figure*}

The crux of course is: Just do it.  We have a dedicated team building Hyperon right now, and we plan to keep going, and to speed up rather than slow down.   A rough overview of the intended development roadmap for the immediately upcoming phase is given in Figure \ref{fig:Hyperon-Roadmap}.

However, this does not exhaust the strategy we have in mind; it will be more effective, we feel, if we can bring in significant additional human resources and insight via developing a robust open source community and strong commercialization efforts.

\subsection{Scaling Up the Hyperon Development Community}

Ben Goertzel summarizes the OpenCog project's history regarding developer relations as follows: {\tt \small ''The name 'OpenCog'was intended to indicate not only the creation of an open-source cognitive architecture (there are many of those of course), but the creation of an open collaborative process among a large community aimed at refining this cognitive architecture toward ambitious goals like AGI at the human level and beyond, as well as toward a variety of practical applications.   However, up till now at least we have pretty much failed at the 'large community' part.   We have done some interesting engineering and research work, have used OpenCog on the back end of a few practical applications, and have pulled in a few brilliant developers from the open source world ... but we have not yet gotten a really sizable active open source community going.   I think this has been for a number of reasons -- OpenCog Classic was never that easy to use in a practical sense, we didn't have simple and accessible wizzy demos for newbie developers to build on, and of course the pursuit of AGI was just not a very popular thing until recently.   It's a whole new era now and AGI is far more interesting to a larger body of developers, so I think we have a strong chance to do with Hyperon some of the things we failed to do with OpenCog Classic, in the development-community domain as well as in the AGI R\&D domain.   But to do this, even in the current climate, we will need to pay sufficient attention to usability and to having easily demonstrable, exciting interim products for devs to play with  -- both of which are things that take edffort, somewhat distinct from the specific effort needed to simply use Hyperon for AGI R\&D among a team of dedicated experts.''}

So far, OpenCog Hyperon has primarily been the work of a small group of developers associated with SingularityNET and True AGI.  The work has frankly been conducted in a somewhat insular manner, although the resulting code has been made open source. 

However, we have a definite plan to open things up development-community-wise as well as code-wise.   The current development roadmap aims for an alpha version of the Hyperon system -- chief components being the MeTTa interpreter and Distributed Atomspace -- around the end of Q1 2024.   Upon this release we will begin a significant outreach initiative aimed at building a sizable active Hyperon developer community.   Given the current enthusiasm for AGI, we expect this will be a less challenging endeavor than it has been OpenCog historically.   However, we don't expect it will necessarily be entirely easy, given that so much of the recent enthusiasm for AGI has been specifically directed toward LLMs rather than more heterogeneous cognitive architectures.

\subsection{Potential Early Demonstrations and Applications}

To accompany the alpha release we are working on a few practical demonstrations of the early Hyperon codebase as well.  Some ideas under current consideration include:

\begin{itemize}
\item A Minecraft-playing assistant that is built on Hyperon and uses program synthesis. 
\item Addition of a rewritable active memory to large language models, which we currently use to control dialogue systems behind humanoid robots like Grace, Sophia, and Desdemona.
\item A Hyperon-based framework for symbolic control of generative AI, particularly for image generation, enabling e.g. the flexible combination multiple stable diffusion models to generate novel structures. 
\item Use of PLN in the Rejuve project, to replace the Bayes Network currently used to estimate biological age based on various attributes of an individual
\end{itemize}

Alongside these relatively straightforward applications, we are also experimenting with employing PLN for knowledge discovery in longevity research. In the context of the Rejuve Biotech project, another SingularityNET initiative, we have been studying long-lived flies that have a lifespan five to eight times longer than regular flies. We have analyzed their DNA and understood, to some extent, why they live longer. We are now interested in applying transfer learning to determine which of these longevity factors can be extrapolated to humans. This is an exciting use case for PLN, particularly as the limited amount of data available does not lend itself to typical DL or ML algorithms in the manner they are normally applied in genomics.

\begin{itemize}
\item Looking a little further forward towards the medium term (approximately the next one to two years), we have two primary focus areas. 
\item Enhancing large language models with neural-symbolic integration to create more intelligent chat systems.   Specifically , as discussed above, we believe coupling Hyperon with LLMs can help overcome the latter's shortcomings in terms of complex multi-step reasoning and fundamental creativity.
\end{itemize}

Controlling a multitude of virtual agents in a video game world, which serves as an excellent testbed for self-modifying codebases.  One likely target here is the use of Hyperon to control an ''Alife species'' called Neoteric in the Sophiaverse metaverse.

\subsection{Commercialization}

While Hyperon is an open-source project, integrated with decentralized infrastructure tools and created with a primary motive of bringing beneficial AGI to the world for the good of all sentient beings (or coming as close to this noble aspiration as possible!)... we do not consider any of this contradictory to the pursuit of practical commercial applications leveraging Hyperon.  To the contrary, we are well aware of the large role that commercial development has played in building out the great open software networks in the world today, e.g. the Linux operating system and the Internet itself.  What is important is that the open-source community has enough vibrancy and energy to it that it operates in close coupling with commercial users and developers, rather than development being entirely driven by the commercial world.  In the modern tech economy, it has been the interplay between the commercial world and the pure R\&D world that has led to some of the most amazing achievements.

With this in mind, we note that several of the authors of this document are involved in a company, True AGI, whose core goal is to use Hyperon to meet the AI needs of enterprises across various vertical markets.   There is also a sister company Zarqa that aims to enhance LLMs using Hyperon, in an attempt to improve commercial ChatGPT-like systems. Other commercial projects mentioned above in passing include Rejuve Biotech, which utilizes Hyperon, among other AI tools, for genomic inference, and Sophiaverse, which employs Hyperion for the operation of metaverse agents.   These projects are all connected with the SingularityNET ecosystem but as Hyperon grows and expands we fully expect it to be leveraged for commercial application by a variety of other parties.

We do recognize that, although commercial projects like these are valuable to generate funds for developers and machine time as well as to build the usable, scalable applications that interface deep AI tools with peoples' everyday lives, they also come with complex challenges.   

There are technical and design challenges.   As Robert Werko, TrueAGI COO, points out, {\tt \small "For commercial development one has to worry about a number of things that are less critical for a research software system, including more rigorous feature prioritization, product-market fit, gathering of user feedback, ease of user onboarding, quality of user experience, security and compliance, and so forth.   Addressing these issues effectively can distract attention from achieving R\&D goals; on the other hand, creating a robustly commerce-ready software framework can enhance one's ability to efficiently and scalably do the R\&D as well."}
 
And, perhaps even more difficult, there are also human and ethical challenges.   A commercial company cannot be solely motivated by a goal like beneficial AGI, it will always at least have a mixed motivation, of working toward the benefit of its owners or shareholders.   However, as Ben Goertzel has put it, {\tt \small "We believe that the ethical pursuit of commercial applications of early-stage AGI system can actually be to the benefit of these AGI system in a moral sense as well as in other dimensions.   One of the key things an AGI needs to learn as it grows up is precisely how to balance its ethical intuitions with the practical activities undertaken by an agent in everyday life in order to get real things done.''}

Initial exploration of commercialization avenues for Hyperon will be carried out, non-exclusively, in collaboration with a handful of SingularityNET ecosystem companies who have been experimenting with OpenCog approaches for some time now.  These include a number of initiatives related to humanoid robots and avatars, including Hanson Robotics with the Sophia robot and others, Awakening Health with the Grace eldercare robot, and Jam Galaxy and Musaic with a line of AI-based music initiatives including Desdemona rock-star bot.   

The Grace robot is currently controlled by a complex dialogue system involving a combination of general-purpose LLMs with custom prompts, specially trained transformer neural nets, and use of the OpenCog Classic system for handling aspects of semantics and memory.   Work is underway to transition from OpenCog Classic to Hyperon here.    The convergence of various sensory inputs with language, cognition and action that is needed for humanoid robots is a demanding but ideal use case for the Atomspace's integrative capabilities.   

A new project called Mind Children, involving roughly meter-tall humanoid robots, is planning to leverage a software architecture similar to Grace's but based on Hyperon from the start, and will focus more heavily on movement and action planning and interaction with physical objects.

The Rejuve Biotech and Rejuve Network projects are already leveraging Hyperon for aspects of bioinformatics data analytics (e.g. the Flybase ontology has been used as an initial test case for the Distributed Atomspace, partly due to its value for analyzing genomics data from long-lived fruit flies for Rejuve Biotech).   As Rejuve Biotech's AI lead Michael Duncan notes,

{\tt \small There are a number of clear reasons why bio-AI needs more than LLMs:

\begin{itemize}
\item Existing biases in research literature (e.g. overrepresentation of a small proportion of human genes and other similar biases; but also more general constraints of existing scientific paradigms) translates into biased training data.  Basically this means LLMs keep recycling our assumptions back to us rather than guiding us in breaking fundamentally new ground -- which is clearly needed given how little we understand about biological systems like human bodies to date.
\item Lack of ability to construct problem relevant numerical simulations.  Answering questions and identifying patterns is only part of the story where bio-AI is concerned, it's also necessary to design and run simulations, which involves complex multi-step reasoning of the sort LLMs are bad at.
\item The well known hallucination problem
\item LLMs lack the critical intuitions that can only be provided by grounding of linguistic biological concepts in biological realities -- such as lab equipment and datasets
\end{itemize}
}

\subsection{Achieving Beneficial AGI}

The potential risks and benefits of AGI have become a highly current and contentious topic since the launch of ChatGPT.   Many of us involved with the Hyperon project, however, have been thinking deeply about these topics for decades, and this thinking has infused itself deeply into the system architecture in ways that go far beyond the generally shallow considerations one sees in popular media.

Ben Goertzel summarizes some of his own views related to AGI ethics, which have been critical for shaping various aspects of the Hyperon project, as follows: 

{\tt \small The various contributors to Hyperon hold a diversity of views on issues related to AGI ethics; there is no rigid ''party line.''  However, a few core hypotheses held by a substantial number of early Hyperon contributors including myself, and thus to some measure underlying the ''initial Hyperon approach'' to AGI ethics are as follows:

\begin{itemize}
\item There are unlikely to be strong guarantees regarding the future consequences of any technology as radical in its nature and broad in its utility as AGI
\item There are unlikely to be strong guaranteed regarding the future consequences of nanotechnology, biotechnology, brain-computer interfacing and a variety of other technologies under current active development around the planet (quite separately from the intersection of these technologies with AGI, which is also a significant point)?\item It seems not plausible that humanity will pause or drastically slow down development of AGI or other advanced technologies, for multiple reasons: e.g. they provide obvious short-term economic and human benefit, and any nation that chose to slow them down would rapidly find itself at dramatic economic and military disadvantage to other nations that didn't make this choice.
\item There is no empirical or logical reason to believe that negative outcomes from the advent of human-level or superhuman AGI are highly likely (argument by reference to Hollywood movies should not be considered compelling; and arguments like that of Nick Bostrom in Superintelligence are barely more so, as Ben Goertzel has argued in detail some years ago \cite{goertzel2015superintelligence}.)
\item Just as human mind/brains are very unlikely to be the most intelligent possible systems realizable using known physics, similarly they are very unlikely to be the most compassionate possible systems realizable.   ''Artificial supercompassion'' is likely just as possible as ''artificial superintelligence.''
\item The various dynamics that make human beings dangerous to each other (and often increasingly dangerous as they achieve more power) are not all necessary aspects of human-level intelligence.   Some of them -- e.g. anger, jealousy, egocentricity -- may be, in large, part, just the way humans happened to evolve.
\item It seems likely possible to create AGI cognitive architectures that are strongly oriented toward compassionate, stably ethical treatment of other sentient beings
\item It seems intuitive that putting early-stage AGI systems to work doing compassionate and beneficial things (e.g. healthcare, education, open science and art) will bias these AGI systems' minds toward compassionate attitudes toward humans and other sentient beings
\item On the whole, the likelihood of positive ethical outcomes appear lower if the first powerful AGIs are controlled by a small group of individuals (e.g. the leadership of one nation or corporation), rather than by a broader swath of humanity.   (This is in part just the maxim that among humans ''power corrupts, and absolute power corrupts absolutely.'').  From this derives the strong value of decentralized infrastructure for AGI, if deployed and utilized correctly (such as e.g. that being developed by SingularityNET and its ecosystem projects like NuNet and Hypercycle)
\item The particulars of current standard human ethical systems, while they may not be natural to AGI systems in the same sense that they're natural to humans, will not be particularly hard for human-level AGIs to master and cope with ... in fact current LLMs, while far short of human-like AGI in various ways, are already quite good at predicting human ethical judgments
\end{itemize}
}

\noindent All these points except the latter have been considered extensively in Goertzel's publications from prior years and decades, e.g. \cite{goertzel2016infusing} \cite{goertzel2015superintelligence} \cite{GoertzelPitt2012}.   

The latter point has been covered in a couple recent articles reporting experiments with GPT4, Llama and other LLMs, specifically examining their ability to predict ethical decisions in various scenarios.  What we found was that LLMs are incredibly proficient at this task, outperforming most individuals in predicting what highly ethical and benevolent humans would do. 

As Ben Goertzel noted in his AGI-23 talk, {\tt \small "These experiments led me to realize that the main issue we face as humans, ethics-wise, is not a lack of knowledge about what the morally right thing to do is, but rather our tendency to prioritize self-interest or tribal interests over ethical considerations. While LLMs lack agency and morality themselves, they possess the capacity to predict ethical decisions based on common human sense.  Now, they can also generate plausible excuses for unethical actions, if you ask them to.   And they still have some work to do, to be able to deal adequately with adversarial attempts to manipulate them to misunderstand situations and suggest or take unethical actions based on this misunderstanding.      But in any case, the base knowledge of what default current human ethics says, is something that's not going to be a problem for any AGI system that is able to effectively leverage LLMs as part of its cognitive process.

"Overall, it is now clear that the hardest part of AGI ethics lies not in AI's understanding of human ethics but rather in dealing with adversarial gamesmanship aimed at inducing misunderstanding, and balancing ethical knowledge with the pragmatic aspects of daily decision-making.   Contrary to the beliefs many have expressed that human ethics is too complex and contradictory for formalization, LLMs have shown the ability to capture these nuances in their own unique way.  

"One conclusion from this is that in the development of goal-oriented cognitive architectures for AGI, employing LLMs as human ethics oracles can quite likely be very valuable. However, it will crucial to employ tools like PLN to verify LLM responses, otherwise adversarial attempts to manipulate them will severely undermine their reliability. " }

In the context of these comments, it would seem the hybrid, multi-component of Hyperon may possibly turn out to be a major asset for AGI ethics.   One can give a Hyperon system a ''brain lobe'' consisting of an LLM fine-tuned to emulate human commonsense ethical judgments, and then when it considers an action for execution, it can use this lobe to assess the ethical valance of this action according to common human standards.   It can also use ethical positivity according to common human standards, as evaluated via this lobe, as one of its top-level goals.

As in human development, the inculcation of appropriate ethics and values in Hyperon systems will be a process and not something to be achieved rigidly once and for all.   It will be a process of, among many other things,

\begin{itemize}
\item incorporating human ethical judgments appropriately into Hyperon system's goals
\item encouraging Hyperon systems to explore and develop their own values in the vicinity of broad universal values like compassion, joy, growth and choice
\item making sure that on balance the applications that early-stage Hyperon systems are carrying out are positive ones that shape its mind in beneficial directions, and help  people in their own quests to guide their own minds in positive directions
\end{itemize}

\noindent There is much to be learned here, and the spirit and ethics of the groups of humans carrying out the work involved will surely be among the most important factors.

\section{Concluding Remarks}

It appears to us that we have an extraordinary opportunity in the next (let's say) 3-10 years to achieve human-level AGI. While the exact timing is uncertain, it seems at least plausible that building a large-scale Hyperon system and allowing it to learn via interacting with it in a suitable variety of application areas could lead to human-level general intelligence. 

And as we have reviewed above, one thing we can plausibly teach a Hyperon system to do is design and write software code.   LLMs are already passable at this in simple contexts; Hyperon is designed to augment this capability with deeper creativity and more capable multi-stage reasoning.   Once we have a system that can design and write code well enough to improve upon itself and write subsequent versions, we enter a realm that could lead to a full-on intelligence explosion and Technological Singularity. 

While we are not yet at a point of having an AGI capable of rapidly self-modifying itself toward Singularity, nevertheless, Hyperon development is currently at a very interesting stage.  We have preliminary versions of a MeTTa interpreter and Distributed Atomspace, enabling experimentation and development.   And as we've noted, alpha versions of these are currently tentatively scheduled for the end of Q1 2024.   

While the system's scalability is still a work in progress, and comprehensive tools and documentation are lacking, these are all on the roadmap and coming fairly soon.   Upon launch of the alpha, we are extremely eager to involve more individuals in the open-source community, whether they are experienced AGI researchers or dedicated AI or application developers.  The potential for rapid advancement toward true, beneficial AGI is enormous.

Hyperon and CogPrime are complex designs and fully understanding them is a work in progress for all of us.   Our aim here has been to outline the highlights, and hopefully to pique the reader's interest in exploring more deeply, for example in the documents, videos and code linked from \url{http//hyperon.opencog.org}.



\bibliographystyle{alpha}
\bibliography{bbm}

\end{document}